\documentclass[10pt,twocolumn,letterpaper]{article}

\usepackage[pagenumbers]{cvpr} % To force page numbers, e.g. for an arXiv version

\definecolor{cvprblue}{rgb}{0.21,0.49,0.74}
\usepackage[pagebackref,breaklinks,colorlinks,allcolors=cvprblue]{hyperref}
\usepackage[bottom]{footmisc}
\usepackage{subcaption}
\usepackage{listings}
\usepackage{upquote}
\usepackage{tabularx}
\usepackage[table]{xcolor}
\definecolor{codegray}{rgb}{0.5,0.5,0.5}
\definecolor{codepurple}{rgb}{0.58,0,0.82}
\definecolor{backcolour}{rgb}{0.95,0.95,0.95}
\lstdefinestyle{mystyle}{
    literate={'}{{'}}1,
    backgroundcolor=\color{backcolour},
    commentstyle=\color{codegray},
    keywordstyle=\color{blue},
    numberstyle=\tiny\color{codegray},
    stringstyle=\color{codepurple},
    basicstyle=\ttfamily\footnotesize,
    breaklines=true,                  % automatic line breaking
    breakatwhitespace=true,
    showstringspaces=false,
    numbers=left,
    numbersep=5pt,
    captionpos=b
}
\def\paperID{*****} % *** Enter the Paper ID here
\def\confName{CVPR}
\def\confYear{2025}

\title{\stb: A Benchmark for Visual Social Inference}

\author{
Neha Balamurugan \qquad Sarah Wu \qquad Adam Chun \qquad \\ Gabe Gaw \qquad Cristobal Eyzaguirre \qquad Tobias Gerstenberg\\
Stanford University \\
{\tt\small nbalamur@stanford.edu \quad gerstenberg@stanford.edu}
}

\usepackage{multirow}
\newcommand{\stb}{\textsc{Spot the Ball}\xspace}

\begin{document}
\maketitle
\begin{abstract}
Humans excel at \textbf{visual social inference}, the ability to infer hidden elements of a scene from subtle behavioral cues such as other people's gaze, pose, and orientation. This capacity drives everyday social reasoning in humans and is critical for developing more human-like AI agents. We introduce \stb, a challenging benchmark for evaluating visual social inference in vision–language models (VLMs) using sports as a test domain. The task is to localize a removed sports ball from soccer, basketball, and volleyball images.
We present a curated evaluation set with human baselines and a scalable pipeline for generating additional test items. We evaluate four state-of-the-art VLMs (Gemini, GPT, LLaMA, Qwen) using three prompting strategies, finding that humans are consistently two to three times more accurate ($20$–$34\%$) than models ($\leq17\%$) across all sports. 
Our analyses show that models rely on superficial spatial heuristics---such as guessing near the image center or nearby players---while humans leverage social cues like gaze direction and body pose. 
These findings reveal a persistent human–model gap in visual social reasoning and underscore the need for architectures that explicitly encode structured behavioral cues to achieve robust, human-like inference.
\end{abstract}
\begin{figure}[t]
        \centering
        \includegraphics[width=0.9\linewidth]{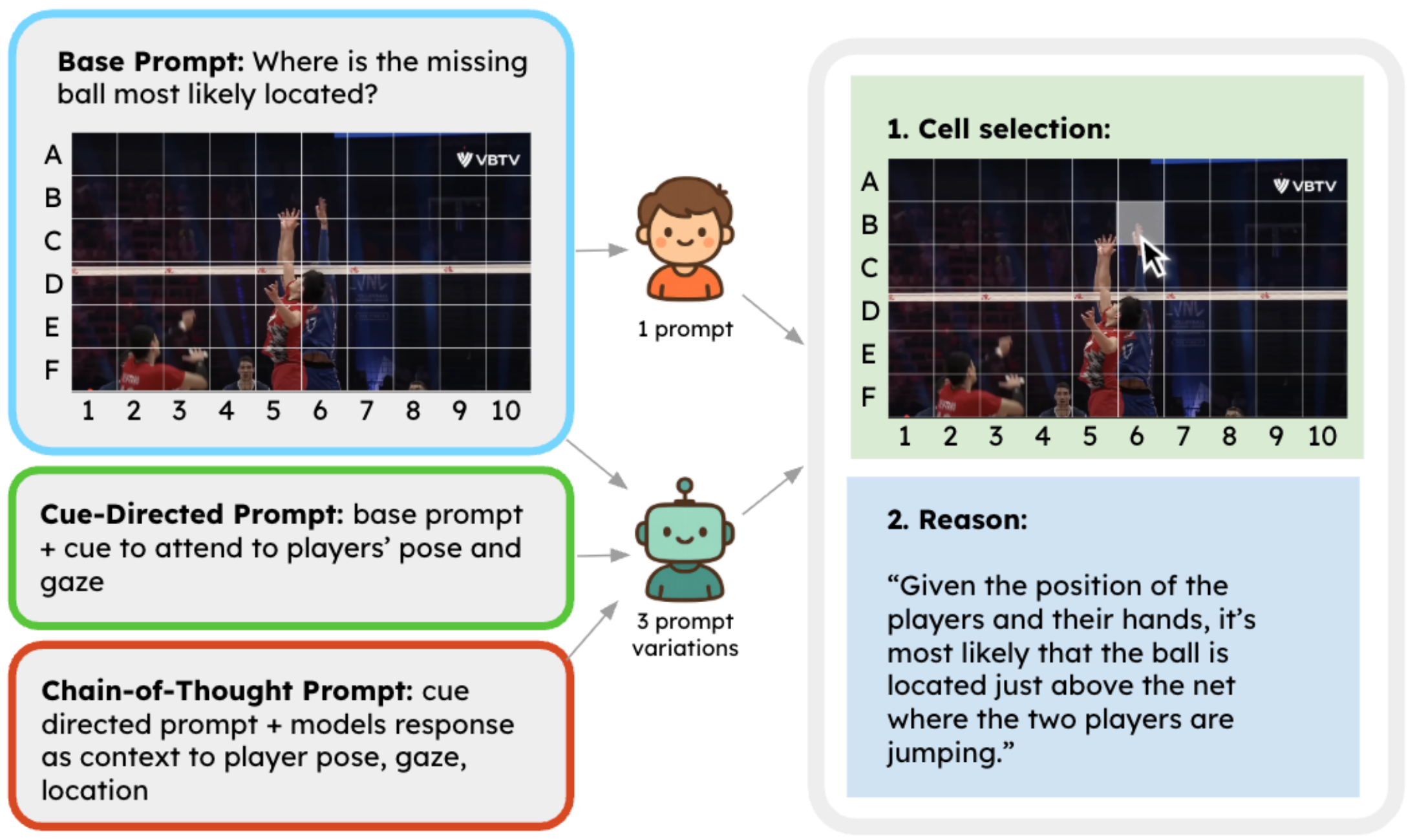}
        \caption{Overview of the \stb task. Given an image with the ball removed, humans and models infer the likely location by reasoning about player pose and gaze. Models are prompted under three conditions, whereas humans receive only the base prompt.}
        \label{fig:intro}
    \end{figure}
%==================================================================

\section{Introduction}
\label{sec:intro}
When someone scans the floor with narrowed eyes, we infer they must be searching for something. When a friend approaches us with open arms, we anticipate a hug. 
As humans, we readily use subtle behavioral cues, such as gaze, pose, and orientation to infer implicit information. 
This ability is rooted in our \emph{theory of mind} (ToM), the capacity to reason about others' beliefs, desires, and intentions to predict their behavior \citep{byom2013theory,dennett1989intentional,wellman2014making}. ToM is fundamental to everyday human interaction \citep{BaronCohen1985,FletcherWatson2014,TagerFlusberg2007}, and increasingly critical for AI systems deployed in social contexts: a robot nurse that misinterprets a patient's gesture, or an autonomous car that fails to anticipate a pedestrian's intention, may behave unsafely.

\begin{figure*}[t]
    \centering
    \includegraphics[width=0.9\linewidth]{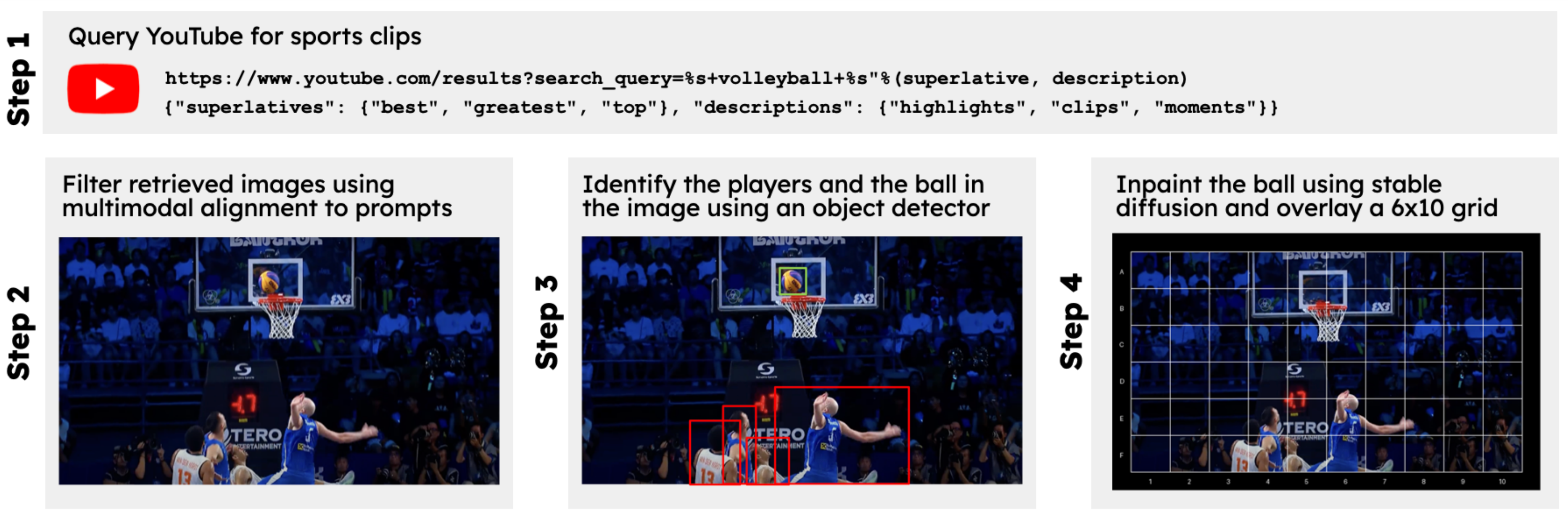}
    \caption{\textbf{Pipeline for constructing the \stb dataset}. We retrieve and filter sports footage from YouTube by alignment to the prompts, detect players and balls using an object detector, and inpaint the ball region with stable diffusion before overlaying a 6×10 grid for location annotation.}
    \label{fig:grid-example}
\end{figure*}

While LLMs perform some types of social reasoning based on text relatively well \citep{gandhi2023understanding}, such behavior may rely primarily on linguistic pattern matching rather than perceptual grounding. Further, human social reasoning rarely occurs through text alone---we continuously integrate visual cues such as gaze direction, facial expressions, body language, and spatial relationships alongside (or in the absence of) verbal information. To evaluate whether AI systems can perform similar visually-grounded social inference, we study VLMs. VLMs process the visual information available to human observers and are increasingly deployed in embodied, safety-critical contexts where robust interpretation of visual social cues is essential.

Existing benchmarks for social reasoning either evaluate models on fully visible scenes \citep{mathur2025socialgenomegroundedsocial,xu2025socialmazebenchmarkevaluatingsocial} or emphasize inanimate objects \citep{liu2025visiblebenchmarkingocclusionperception}, failing to capture how humans reason about others under partial information. We introduce \stb, inspired by a classic newspaper puzzle, to evaluate whether VLMs can infer hidden objects from visual social cues in real-world images. Ball sports provide an ideal testbed as players' gaze, posture, and positioning are causally coupled to the ball, yielding interpretable social signals. We use static images to isolate social reasoning from motion dynamics.

We contribute: (1) 150 curated sports images with human baselines, (2) systematic evaluation of four leading VLMs under multiple prompting strategies, and (3) a scalable pipeline for generating ball sports reasoning tasks, with which we have produced 3,000 additional soccer images for training and analysis. We provide the first structured evaluation of VLMs' ability to leverage social cues for inferring hidden objects in real-world scenes. Our results reveal large, consistent gaps between human and model performance.

%==================================================================

\section{Related Work}

\paragraph{Human social reasoning.}
Humans acquire social reasoning abilities early in life, learning from infancy to represent and infer others' beliefs, desires, and intentions \citep{wellman2014making,sodian2011theory,wang2025modeling}. Adults readily infer rich social states from minimal visual cues---moving geometric shapes \citep{heider1944experimental, tremoulet2000perception}, gaze \citep{kleinke1986gaze}, facial expressions \citep{todorov2015social}, and body language \citep{degelder2015perception}. As AI systems are increasingly deployed in social contexts, they must exhibit similar reasoning capabilities for seamless human interaction \citep{mathur-etal-2024-advancing}.

\paragraph{Social reasoning benchmarks for AI.}
Most social reasoning benchmarks for AI operate in text-only settings, evaluating ToM \citep{hitom,tomchallenges,gandhi2023understanding}, empathy \citep{chen2024emotionqueenbenchmarkevaluatingempathy,apathy}, moral reasoning \citep{morables,moralchoice}, deception detection \citep{chen2024deceptiondetectiondeeperdataset,drex}, and negotiation \citep{negotiation,sap2019socialiqacommonsensereasoningsocial}. While informative, these benchmarks cannot assess whether models extract social cues from visual perception---the primary modality through which humans perform social inference.

Video-based benchmarks extend social reasoning evaluation to dynamic visual contexts, including theory-of-mind in human interactions \citep{mmtomqa,mumatom,li2025egotombenchmarkingtheorymind} and synthetic simulations \citep{fan2025somitomevaluatingmultiperspectivetheory}, causal and counterfactual video QA \citep{xiao2021nextqanextphasequestionansweringexplaining,wu2023acquireddatasetansweringcounterfactual}, and social interaction understanding \citep{socialiq1,wilf2023social}. Static image benchmarks are less common but include Visual Commonsense Reasoning \citep{zellers2019recognitioncognitionvisualcommonsense}, facial expression understanding \citep{affectnet,emotic}, and social relationship recognition \citep{li2018visualsocialrelationshiprecognition}. However, existing visual benchmarks either present fully observable scenes \citep{mathur2025socialgenomegroundedsocial,xu2025socialmazebenchmarkevaluatingsocial} or focus on physical occlusion without social cues \citep{liu2025visiblebenchmarkingocclusionperception,pothiraj2025}. No prior work evaluates whether models can infer hidden information purely from visual social cues.

\paragraph{Our contribution.}
We introduce a zero-shot, image-based benchmark requiring models to infer a hidden object by integrating pose, gaze, and orientation cues from human agents. Unlike prior work, our task isolates visually-grounded social reasoning in naturalistic scenes with partial information, reflecting how humans actually perform social inference in everyday contexts.

%==================================================================

\begin{figure*}[t]
    \centering
    \includegraphics[width=1\linewidth]{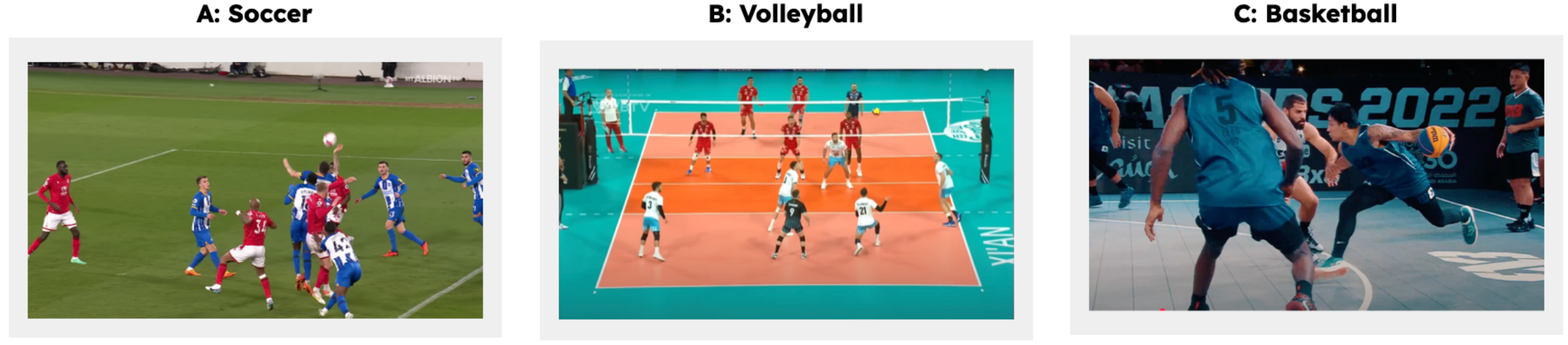}
    \caption{\textbf{Player density and coverage across sports in the \stb dataset.} In Soccer frames (A) player count and area are intermediate compared to the other sports. Volleyball frames (B) feature the most players but each occupies a smaller visual area, providing weaker pose and gaze cues. Basketball (C) has fewer yet larger players, offering clearer postural and gaze information. }
    \label{fig:both_wball}
\end{figure*}

\section{\stb}
% \subsection{Task Overview}
In \stb, the objective is to infer the location of a removed ball in a sports frame (see \autoref{fig:intro}). This task evaluates a model's ability to localize a hidden object through reasoning over social and physical contextual cues such as players' gaze, body orientation, and spatial positioning, rather than relying on direct visual evidence of the object itself in addition to sport specific knowledge.

\subsection{Evaluation Set}
We curated 150 images from publicly available soccer, basketball, and volleyball footage on YouTube (see \autoref{fig:grid-example}). Frames were procedurally selected to maximize contextual informativeness (non-occluded players, clear ball presence, spatial distribution) and manually verified for quality. The ball was removed via inpainting after recording its ground truth location. This evaluation set is used in our experiments for comparing models against human reasoning.

\subsection{Scalable Pipeline}
To scale beyond the evaluation set, we developed a modular pipeline to generate realistic inpainted sports scenes: 
\begin{enumerate}
\item \textbf{Video retrieval.} 
    Broadcast footage was retrieved from YouTube using sport-specific queries with action-focused keywords (``best'', ``highlights'', ``moments''). Videos were decoded with OpenCV \citep{opencv_library} and sampled at $\sim$1 FPS.
    
    \item \textbf{Frame filtering.}
    Each frame was scored with CLIP \citep{radford2021learningtransferablevisualmodels} against prompts like ``picture of volleyball players in action with ball''. Only frames exceeding a similarity threshold were retained.
    
    \item \textbf{Ball and player detection.} 
    Frames were passed through YOLOv8 \citep{yolov8_ultralytics} to detect balls and players. We filtered by confidence and spatial plausibility, requiring exactly one ball per frame, non-overlapping with and proximal to players. This eliminated spurious detections while preserving contextual cues.
    
    \item \textbf{Ball inpainting.} 
    Ball regions were removed and filled using Stable Diffusion inpainting, which generates realistic textures and lighting while avoiding visible artifacts. Player masks ensured body posture and gaze cues remained intact. Images were manually checked to remove any remaining ball shadows or artifacts.
    
\end{enumerate}

Each image is overlaid with a $6 \times 10$ alphanumeric grid (rows A--F, columns 1--10). Ground-truth labels correspond to coordinates covering the original ball location (e.g., \texttt{[A5]} for single-cell coverage, or \texttt{[A5, A6, B5, B6]} for multi-cell overlap).

This pipeline generated 3,000 additional soccer images beyond the curated evaluation set. The modular design allows extensions to other ball sports or difficulty controls (e.g., varying player density or occlusion severity).

%==================================================================

\section{Experiments}
To assess visual social inference in models and humans, we test four VLMs using three prompting strategies under three sports in the \stb task and compare them to human performance. The participants, both humans and models, select one grid cell (e.g., ``B6'') in addition to a text reasoning. Predictions are evaluated against the ground-truth set of valid cells. Multiple adjacent cells may be considered correct if they overlap with the ball region. Then, we compare model and human performance across several quantitative and behavioral metrics to identify accuracy trends and reasoning patterns. 

\subsection{Models}  
We evaluate four multi-modal instruction-following models: Gemini-2.0-flash-001, GPT-4.1-mini, LLaMA-3.2-11B-Vision-Instruct and Qwen-2.5-VL-7B-Instruct. This set spans both closed and open-weight paradigms, all supporting high-resolution vision inputs and free-form text reasoning. 

While the specifics of the proprietary models are not publicly disclosed, available evidence suggests that both Gemini-2.0-flash-001 and GPT-4.1-mini adopt unified transformer backbones that fuse visual and textual representations through shared cross-modal attention layers. LLaMA-3.2-Vision-Instruct and Qwen-2.5-VL-7B-Instruct both pair pretrained language models with Vision Transformer–based encoders that extract image features and integrate them into the text model using adapters to enable reasoning across modalities. Qwen-2.5 handles native-resolution inputs by using convolutional and windowed-attention blocks.\citep{bai2025qwen25vltechnicalreport}. Together, these architectures represent a spectrum from fully integrated multimodal transformers to adapter-based and hierarchical fusion strategies. 

\subsection{Domains}  
We selected soccer, volleyball, and basketball because they are ball sports that are present in high-frequency in pretraining corpora, allowing the models to have an understanding of the game mechanics. Further, these sports differ from each other uniquely in how many players they contain (we use clips of 3v3 basketball) and how long the ball can be with a player and these differences in the mechanics lead to the footage of these sports having variations in the amount of information and visual density. This variation allows us to analyze how models generalize under different types of visual ambiguity which might be relevant to model downfalls (\autoref{tab:sport_comparison}, \autoref{fig:both_wball}).  

\begin{table}[b]
    \centering
    \caption{Statistics across 3 sports in the evaluation set.}
    \label{tab:sport_comparison}
    \resizebox{\columnwidth}{!}{%
    \begin{tabular}{lccc}
        \toprule
        & Soccer & Volleyball & Basketball \\
        \midrule
        Avg. grid cells ball spans & 2.27 & 2.08 & 2.42 \\ 
        Avg. ball pixel area & 785.37 & 904.86 & 1631.27 \\
        Avg. distance of ball from center & 122.81 & 188.02 & 163.30 \\
        Avg. players in scene & 4.26 & 9.92 & 5.46 \\
        Player coverage (pixels) & 13718.87 & 6600.26 & 20067.43 \\
        \bottomrule
    \end{tabular}
    }
\end{table}

\subsection{Prompting Strategies}  
We test the models on three variations of prompts that are provided in addition to the encoded image:  
\begin{itemize}
    \item \textbf{Base Prompt}: Instruction to provide the cell location of the missing ball.
    \item \textbf{Cue-Directed Prompt}: The Base Prompt with the additional cue to focus on players' pose and gaze.
    \item \textbf{Chain-of-Thought Prompt} (CoT): First, we ask one-shot questions about the players' location, pose and gaze (3 questions total). Next, the responses to these, are provided as \texttt{context} and asked to predict the grid cell. 
\end{itemize}

To estimate distributional behavior, we sample $n=50$ predictions per image at Base Prompt and Cue-Directed Prompt, and $n=20$ at Chain-of-Thought Prompt, all at temperature $T=0.6$ for all models. 
These strategies are motivated by prior findings that CoT improves performance on spatial and visual reasoning~\citep{li2024topviewrsvisionlanguagemodelstopview,chen2024spatialvlmendowingvisionlanguagemodels}, and that auxiliary cues can enhance localization~\citep{pothiraj2025}. Supplementary material contains system prompts.
\begin{figure*}[t]
    \centering
    \includegraphics[width=0.9\linewidth]{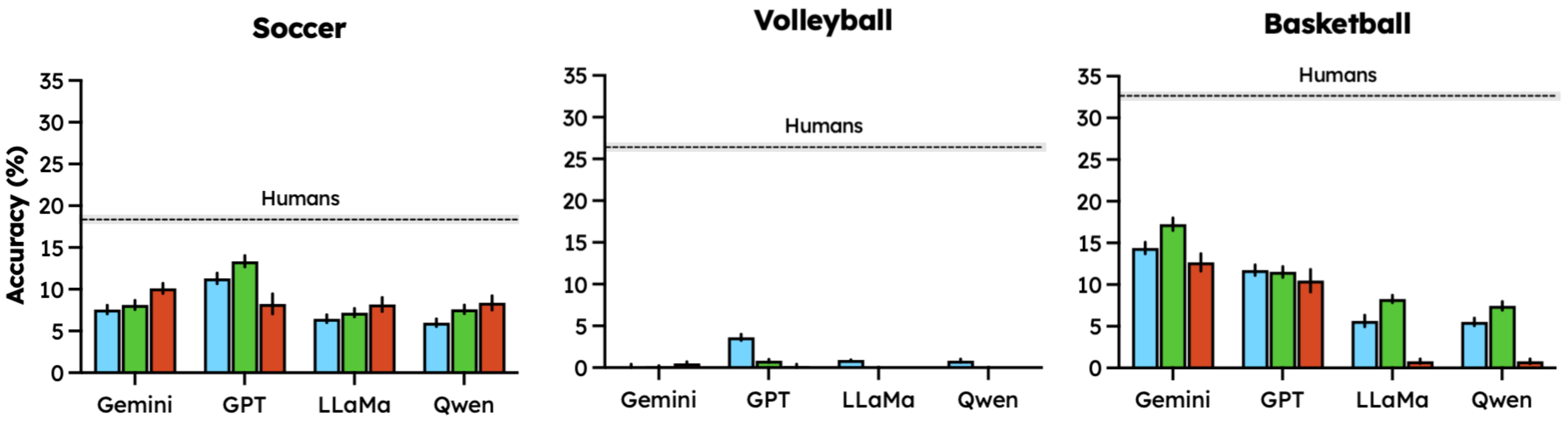}
    \caption{\textbf{Accuracy}. Model accuracy in each sport under different prompting strategies (blue $=$ base prompt, green $=$ cue-directed prompt, red $=$ chain-of-thought prompt). The dashed line shows human accuracy using the base prompt in the given sport. Error bars and gray ribbon show 95\% bootstrapped confidence intervals.}
    \label{fig:accuracy_sport}
\end{figure*}
\subsection{Human experiment}
To collect human baselines, recruited 176 participants from  Prolific and compensated them for their time ($\$12$/hour base + up to $\$1$ accuracy bonus). After excluding 26 for failed attention checks, the final sample was $N=150$ (50 per sport). The experiments were pre-registered on the \href{https://osf.io/a52rw}{Open Science Framework}\footnote{The volleyball and basketball conditions were identical to the preregistered soccer experiment. Model prompts were updated from pre-registration based on piloting.} and approved by the [REDACTED] Institutional Review Board.
Participants received instructions with example images (ball visible and removed) before making three guesses per test image by clicking grid cells. Each participant saw 52 images (50 test, 2 attention checks with visible balls) from one sport in randomized order. Participants who clicked outside ball-containing cells on attention checks were excluded.
Completion time averaged 17.3 minutes ($\text{SD}=7.4$).

\subsection{Evaluation Metrics}
\label{sec:metrics}
We evaluate models across three dimensions: \textbf{task performance} (localization accuracy), \textbf{alignment with humans} (similarity to human reasoning), and \textbf{behavioral strategies} (prediction patterns).

\paragraph{Task performance.}
(1) \textbf{\textit{Accuracy:}} For image \(i\) with ground-truth cells \(\mathcal{G}_i\) and prediction \(\hat{y}_i\):
\begin{equation}
\text{Accuracy} = \frac{1}{N}\sum_{i=1}^N \mathbb{1}\bigl[\hat{y}_i \in \mathcal{G}_i\bigr]. 
\end{equation}

(2) \textbf{\textit{Euclidean Error: }} Mean minimum distance (in pixels) from predicted cell center \(c(\hat{y}_i)\) to nearest ground-truth cell:
\begin{equation}
d_i = \min_{g \in \mathcal{G}_i} \|c(\hat{y}_i) - c(g)\|_2.
\end{equation}

% \paragraph{Task performance}
% \begin{enumerate}
%     \item Accuracy:
% For each image \(i\), let \(\mathcal{G}_i\) be the set of ground‐truth grid cells and \(\hat{y}_i\) the predicted cell. Accuracy is the proportion of predictions landing in any valid cell:

% \begin{equation}
% \text{Accuracy} = \frac{1}{N}\sum_{i=1}^N 1\bigl[\hat{y}_i \in \mathcal{G}_i\bigr]. 
% \end{equation}

% \item Euclidean Error:
% We measure the minimum distance in pixels from the predicted cell center \(c(\hat{y}_i)\) to the nearest ground-truth cell:
% \begin{equation}
% d_i = \min_{g \in \mathcal{G}_i} \|c(\hat{y}_i) - c(g)\|_2.
% \end{equation}

% The reported score is the mean of \(d_i\) across all images.
% \end{enumerate}

\paragraph{Alignment with humans.}
(1) \textbf{\textit{Wasserstein Distance:}} We compare model and human prediction distributions \(P\) and \(Q\) using Earth Mover's Distance:
\begin{equation}
\begin{aligned}
W(P,Q) = \min_{\gamma \geq 0} \sum_{j,k} \gamma_{jk}\,D_{jk} \\
\text{s.t.}\;\; \sum_k \gamma_{jk}=P_j,\;\; \sum_j \gamma_{jk}=Q_k,
\end{aligned}
\end{equation}
where \(D_{jk}\) is Euclidean distance between cell centers. Lower values indicate closer alignment with human strategies.

% \paragraph{Alignment with humans}
% \begin{enumerate}
%     \item Wasserstein Distance:
% To assess whether the distribution of model guesses resembles that of human guesses, we compare their empirical distributions \(P\) and \(Q\) over the $6{\times}10$ grid with the Earth Mover's Distance:

% where \(D_{jk}\) is the Euclidean distance between cell centers \(j,k\). This was calculated using the \texttt{wasserstein\_distance} method from \texttt{scipy} package in python. Lower values indicate distributions more similar to human strategies.
% \end{enumerate}

\paragraph{Behavioral strategies.}
Let $\mathcal{B}_i=\{b_{i,m}\}$ be player bounding boxes in image $i$, with $p_{i,t}=c(\hat y_{i,t})$ the predicted cell center. Define point-to-box distance:
\begin{equation}
\mathrm{dist}(x,b) = 
\begin{cases}
0, & x\in \mathrm{supp}(b),\\
d(x, \mathrm{supp}(b)), & \text{otherwise},
\end{cases}
\end{equation}
where $d(x, \mathrm{supp}(b))$ is Euclidean distance to the box.

(1) \textbf{\textit{Near Player Rate (NR):}} Fraction of predictions within threshold $\tau D$ of any player ($\tau=0.08$, $D=$ image diagonal):
\begin{equation}
NR = \frac{1}{\sum_i T_i}\sum_{i,t}\mathbb{1}\!\left[\,\min_{b\in\mathcal{B}_i}\mathrm{dist}\bigl(p_{i,t},b\bigr) \le \tau D\,\right].
\end{equation}

(2) \textbf{\textit{Overlap Rate (OR):}} Fraction of predictions whose grid cell intersects a player box by $\ge\theta$ of cell area ($\theta=0.02$):
\begin{equation}
OR = \frac{1}{\sum_i T_i}\sum_{i,t}\mathbb{1}\!\left[
\max_{b\in\mathcal{B}_i}\frac{\mathrm{area}\bigl(c(\hat y_{i,t})\cap b\bigr)}{\mathrm{area}\bigl(c(\hat y_{i,t})\bigr)} \ge \theta
\right].
\end{equation}

(3) \textbf{\textit{Center Ratio (CR):}} Ratio of prediction mass to ground-truth mass in central window $\mathcal{S}=\{(r,c): r\in\{2,3,4,\}, c\in\{3,\ldots,7\}\}$ ($3{\times}5$ region):
\begin{equation}
CR = \frac{\sum_{j \in \mathcal{S}} p_j}{\sum_{j \in \mathcal{S}} q_j},
\end{equation}
where $p$ is the model distribution and $q$ is the ground-truth prior. Values $>1$ indicate center bias.

(4) \textbf{\textit{Entropy:}} Normalized entropy measures prediction spread:
\begin{equation}
    \hat H(p) = \frac{-\sum_{j=1}^{60} p_j \log p_j}{\log 60}.
\end{equation}
Higher values indicate broader exploration; lower values indicate concentration.

%==================================================================

\section{Results and Discussion}

\label{sec:results}

\begin{table}[b!]
\centering
\caption{\textbf{Mean Euclidean error} (± std) for humans and models across three sports and prompting types. Base = base prompt, Cue = cue-directed prompt, and CoT = chain-of-thought prompt. Lower scores reflect closer predictions to the true ball location.}
\resizebox{\columnwidth}{!}{
\begin{tabular}{llccc}
    \toprule
    \textbf{Model} & \textbf{Prompts} & \textbf{Soccer} & \textbf{Volleyball} & \textbf{Basketball} \\
    \midrule
    Human  & Base & 113.4$\pm$65.1 & 72.0$\pm$40.1 & 68.5$\pm$40.8 \\
    \midrule
    \multirow{3}{*}{Gemini} & Base & 139.1$\pm$79.2 & 151.9$\pm$54.9 & 132.2$\pm$81.4 \\
           & Cue & 133.3$\pm$75.3 & 151.5$\pm$51.5 & 119.3$\pm$75.7 \\
           & CoT & 141.1$\pm$72.9 & 150.5$\pm$48.2 & 134.7$\pm$73.4 \\
    \midrule
    \multirow{3}{*}{GPT}    & Base & 135.6$\pm$79.4 & 142.7$\pm$58.5 & 127.7$\pm$69.8 \\
           & Cue & 139.6$\pm$88.6 & 148.6$\pm$52.2 & 125.3$\pm$70.2 \\
           & CoT & 146.1$\pm$68.3 & 155.3$\pm$56.8 & 137.1$\pm$68.8 \\
    \midrule
    \multirow{3}{*}{LLaMA}  & Base & 143.3$\pm$79.9 & 172.8$\pm$61.4 & 161.6$\pm$95.3 \\
           & Cue & 147.0$\pm$83.1 & 163.7$\pm$67.2 & 147.2$\pm$90.0 \\
           & CoT & 140.2$\pm$87.1 & 272.6$\pm$50.7 & 211.4$\pm$82.6 \\
    \midrule
    \multirow{3}{*}{Qwen}   & Base & 142.6$\pm$80.2 & 170.9$\pm$60.7 & 162.9$\pm$95.9 \\
           & Cue & 147.6$\pm$81.9 & 162.9$\pm$66.6 & 147.2$\pm$90.7 \\
           & CoT & 139.0$\pm$81.0 & 271.5$\pm$52.9 & 211.0$\pm$82.5 \\
    \bottomrule
\end{tabular}}
\label{tab:euclidean}
\end{table}

We evaluate humans and four VLMs across three sports and three prompting strategies based on accuracy, spatial error, and distributional analyses. The overarching finding is a large human--model gap in both accuracy and approach. Because ball sports are highly structured and heavily represented in web-scale pretraining, a lack of generic world knowledge is an unlikely driver of errors. Instead, we aim to discern if the models fail at either \textbf{identifying} the relevant social cues, \textbf{extracting} them or \textbf{composing} them to localize the location of the missing ball.
\begin{figure*}[t]
    \centering
    \includegraphics[width=0.9\linewidth]{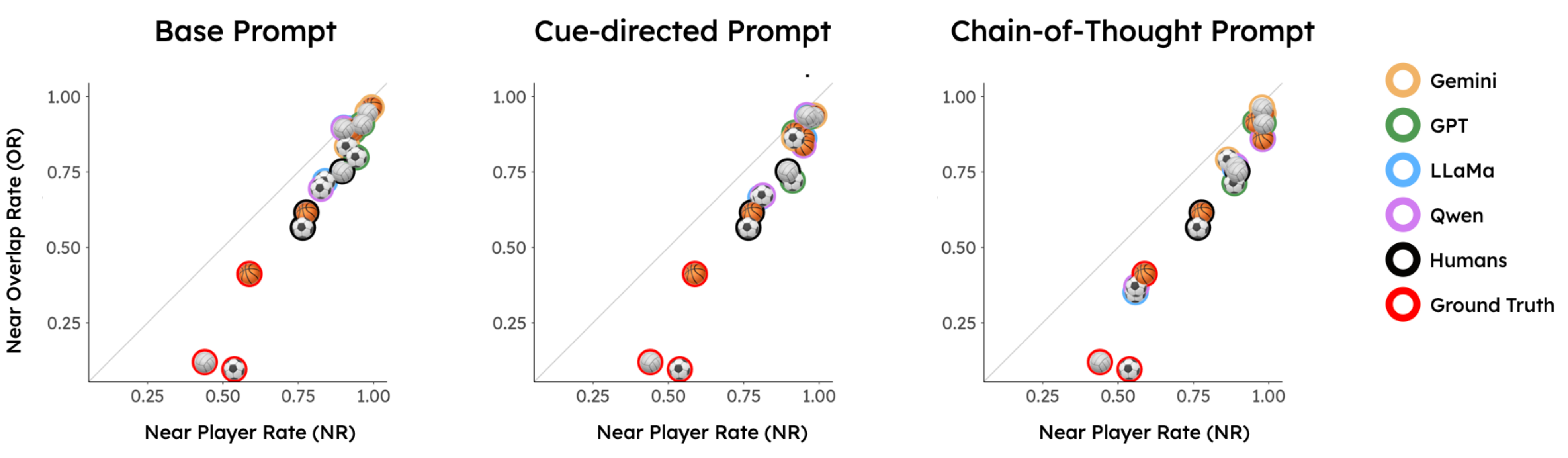}
    \caption{\textbf{Player Proximity Analysis}. Each point corresponds to a model–sport combination. The $x$-axis shows the fraction of guesses within a fixed distance threshold of any player (Near Player Rate), while the $y$-axis shows the fraction of guesses whose predicted cell overlaps a player bounding box (Near Overlap Rate). 52.2\% of ground truth balls are near players, 20.9\% of the ground truth balls are near players by overlap.}
    \label{fig:proximity}
\end{figure*}
\subsection{Quantitative Performance}
\paragraph{Humans outperform all models by a large margin.} 
Humans consistently outperform models in predicting the ball's location. 
Across sports, human accuracy ranges from $19$--$34\%$, while all models remain at or below $17\%$ (see \autoref{fig:accuracy_sport}). 
The accuracy gap is not due to the models being more likely to produce close misses.  The Euclidean errors in \autoref{tab:euclidean} show that model predictions are often far from the true location. The distances from the correct locations are larger for models than for humans. In volleyball, where humans are most precise ($72.0 \pm 40.1$ pixels), the models' error is about twice as large on average.
Moreover, performance does not always improve with richer prompts. In fact, for models like Llama and Qwen, Chain-of-Thought prompts amplify errors in certain cases (e.g., LLaMA reaching $272.6 \pm 50.7$ pixels in volleyball). 

\paragraph{Models and humans find different sports challenging.} 
Performance across sports differs between humans and models. Humans perform best in basketball, worse in volleyball, and worst in soccer, while models perform similarly in basketball and soccer but struggle most in volleyball. This discrepancy suggests that humans and models rely on different visual cues to infer ball location. 

As shown in \autoref{tab:sport_comparison}, basketball scenes feature fewer players (5.5 on average) who occupy the largest proportion of the frame ($\sim$20{,}000 pixels per player), making pose and gaze cues clearer and likely contributing to the highest human accuracy. Volleyball, by contrast, includes nearly twice as many players (9.9 on average) but with much lower per-player pixel density ($\sim$6{,}600), reducing the salience of individual cues. Nevertheless, humans may still aggregate directional information across multiple players, leading to intermediate performance compared to soccer, where both the number of players and their coverage fall in between. 

Models, however, perform poorly in volleyball partly because the ball is rarely in contact with players as it is struck rather than held, making a ``guess-near-player'' heuristic unreliable. As shown in \autoref{fig:proximity}, models (90\%) are more likely than humans (65--75\%) to predict that the ball lies near a player, a bias that fails in volleyball where the ball often travels away from them.

\begin{figure}[b]
\centering    
\includegraphics[width=.8\linewidth]{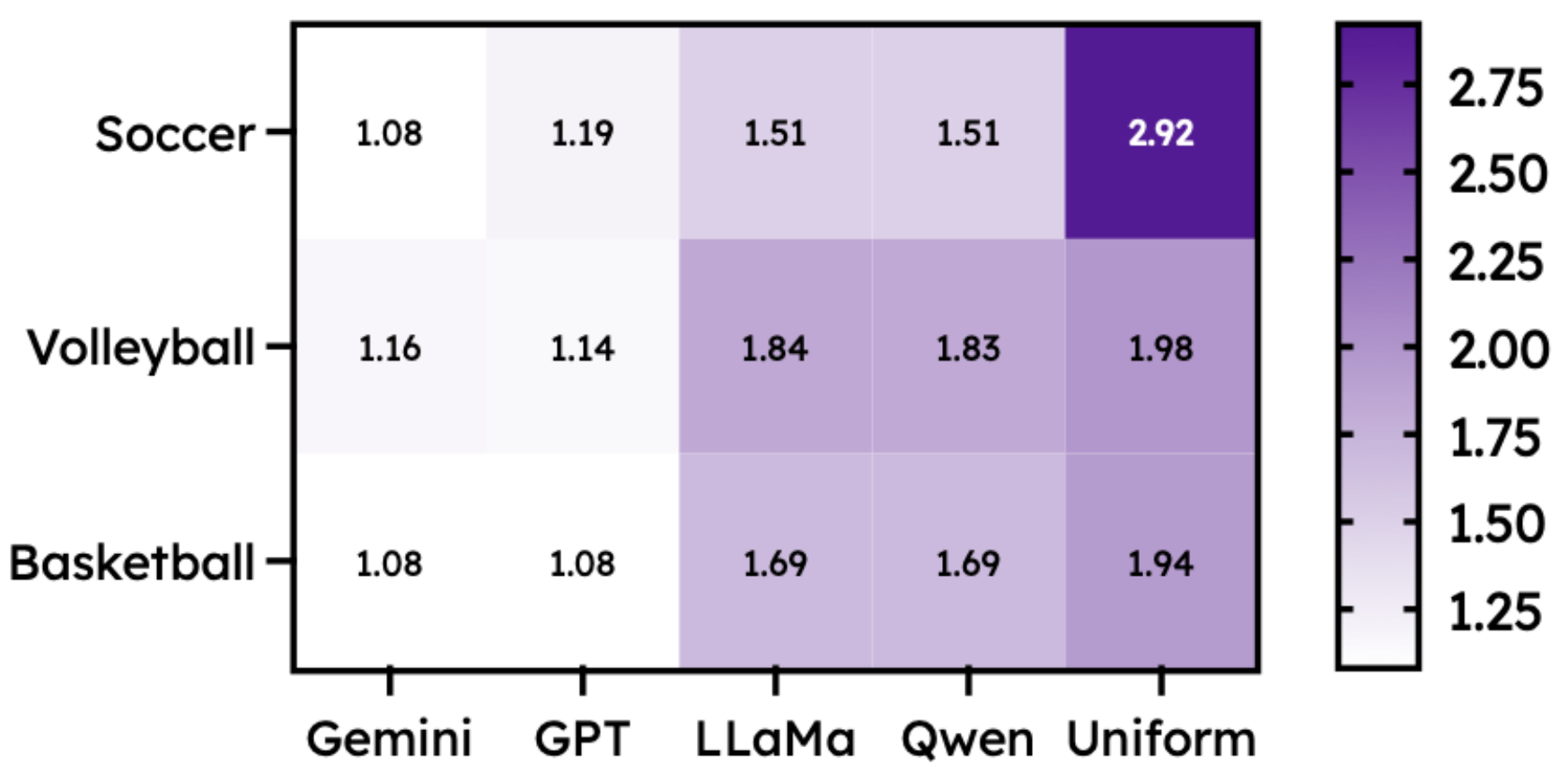}
    \caption{\textbf{Wasserstein Distance}. Calculated between model and human predictions across sports. Lower values indicate distributions closer to human guesses.}
    \label{fig:wasserstein}
\end{figure}

\paragraph{Both model and human guesses are center biased.} 
The percentages of ground truth cells in the center window  are: soccer (36.3\%), volleyball (46.2\%), basketball (56.3\%). As noted in \autoref{tab:centrality_combined}, models exhibit a strong bias toward predicting central grid cells, reflected in elevated center ratios ($R>1.2$). Humans also show a high center bias ($R=1.63$) and exhibit higher normalized entropy ($\hat{H}=0.855$) than all models ($0.698$--$0.808$). 
So even though humans are more likely to guess that the ball is in the center compared to models, their answers also exhibit more variance, too. This suggests that humans consider more possibilities than models do. Models may rely on simpler strategies like ``guess near a player'' or  ``guess near the center'', despite a temperature of 0.6 and repeated sampling.
% Further the higher center ratio in the volleyball case could be demonstrative of the idea that models guess near the central player -- a combination of the two heuristics we proved earlier. 

% \begin{table}[h]
% \centering
% \caption{Center bias ($R$) and normalized entropy ($\hat{H}$) for each model. Humans show higher entropy and broader distributions, while models exhibit stronger central bias and narrower spread.}
% \begin{tabular}{lcc}
%     \toprule
%     \textbf{Model} & \textbf{Center Ratio $R$} & \textbf{Entropy $\hat{H}$} \\
%     \midrule
%     Human  & 1.628 & 0.855 \\
%     Gemini & 1.378 & 0.795 \\
%     GPT    & 1.356 & 0.808 \\
%     LLaMA  & 1.262 & 0.698 \\
%     Qwen   & 1.257 & 0.702 \\
%     \bottomrule
% \end{tabular}
% \label{tab:distribution}
% \end{table}

\paragraph{Models and humans distribute guesses differently.}
\autoref{fig:wasserstein} reveals that the overall structure of the models' predictions diverges from that of humans. The distribution of guesses by open source models (Qwen and Llama) are less similar to human response distributions compared to those of proprietary models (GPT and Gemini). All models perform better than a baseline which predicts a uniform distribution of guesses.  
% As demonstrated earlier with the entropy, humans use cue-driven distributions (as explained by the textual responses) that place probability mass across multiple plausible regions, but model predictions tend to collapse. This produces consistently high Wasserstein distances. Models aren't simply ``noisier'' versions of humans -- they behave systematically differently. 

% \autoref{fig:wasserstein} indicates that model predictions diverge from human response distributions, with open-source models (Qwen, LLaMA) farther from humans than proprietary models (GPT, Gemini); all outperform a uniform baseline. To connect this with our entropy analysis, note that the two metrics capture complementary aspects: entropy measures dispersion (how broadly probability mass is spread), whereas Wasserstein distance measures alignment (how the mass is positioned relative to the human distribution).

Generally, humans exhibit higher-entropy, cue-driven distributions that place probability across multiple plausible regions (as reflected in their textual explanations). Models, by contrast, often collapse mass onto narrower regions, yielding lower entropy and, correspondingly, higher Wasserstein distance from humans. Importantly, entropy also helps understand why Wasserstein is high: (i) when model entropy is much less than human entropy, divergence could stem from under-dispersion of guesses; (ii) when entropies are comparable yet Wasserstein distance remains high, divergence could reflect misplaced mass (e.g. systematic center or near-player bias), not just spread. Taken together, entropy contextualizes whether distributional mismatch is due to how much probability is spread versus where it is placed, clarifying that models follow strategies distinct from humans rather than behaving as merely ``noisier'' variants.

\begin{table}[b]
\centering
\caption{\textbf{Center ratio ($R$) and normalized entropy ($\hat{H}$)} for each model across sports, with aggregate values across all sports. $R>1$ indicates a center bias (meaning, the ball is predicted to be closer to the center than it actually is). Higher $\hat{H}$ indicates broader distribution of predictions.}
\resizebox{\columnwidth}{!}{
\begin{tabular}{llcc}
\toprule
\textbf{Sport} & \textbf{Model} & \textbf{Center Ratio $R$} & \textbf{Norm.\ Entropy $\hat{H}$} \\
\midrule
\multirow{5}{*}{Soccer} &
Gemini & 0.989 & 0.763 \\
& GPT    & 0.732 & 0.792 \\
& LLaMA  & 1.150 & 0.607 \\
& Qwen   & 1.131 & 0.611 \\
 \cmidrule{2-4}
& Human  & 1.164 & 0.817 \\
\midrule
\multirow{5}{*}{Volleyball} &
 Gemini & 1.697 & 0.721 \\
 & GPT    & 1.487 & 0.710 \\
 & LLaMA  & 0.945 & 0.420 \\
 & Qwen   & 0.953 & 0.422 \\
\cmidrule{2-4}
& Human  & 1.602 & 0.768 \\
\midrule
\multirow{5}{*}{Basketball} & 
 Gemini & 0.801 & 0.736 \\
 & GPT    & 0.883 & 0.737 \\
 & LLaMA  & 0.510 & 0.515 \\
 & Qwen   & 0.518 & 0.517 \\
 \cmidrule{2-4}
 & Human  & 1.093 & 0.801 \\
% \midrule
% \multirow{5}{*}{Aggregate} 
%  & Gemini & 1.378 & 0.795 \\
%  & GPT    & 1.356 & 0.808 \\
%  & LLaMA  & 1.262 & 0.698 \\
%  & Qwen   & 1.257 & 0.702 \\
%  \cmidrule{2-4}
%   & Human  & 1.628 & 0.855 \\
\bottomrule
\end{tabular}}
\label{tab:centrality_combined}
\end{table}

\begin{figure*}[t]
    \centering
    \includegraphics[width=.8\linewidth]{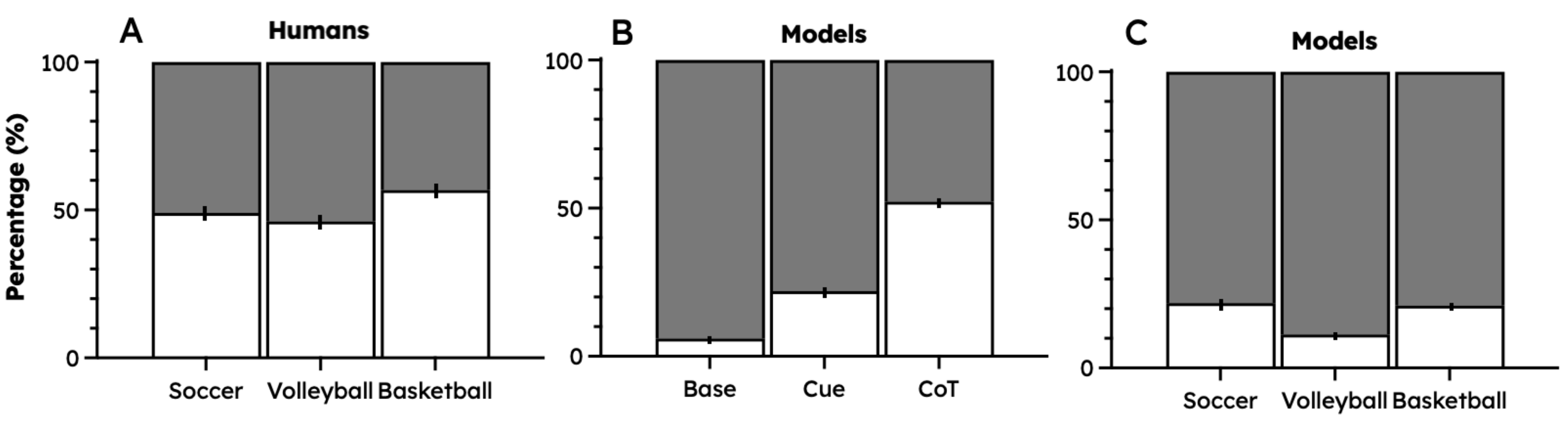}
    \caption{\textbf{Embedding analysis of reasoning similarity.} Each bar shows the proportion of model rationales whose sentence embeddings are closer to pose-like versus gaze-like reasoning templates, separated by sports and prompting strategies. Error bars indicate 95\% bootstrapped confidence interval. % Higher percentages toward one category indicate stronger reliance on that cue type in the model’s explanations.
    The distribution of explanations with higher presence of gaze-like vs. pose-like language is shown for (A) humans separated by sport, (B) models separated by sport, and (C) models separated by prompting strategy.}
    \begingroup
    \phantomsubcaption
    \label{subfig:posevsgaze_a}
    \phantomsubcaption
    \label{subfig:posevsgaze_b}
    \phantomsubcaption
    \label{subfig:posevsgaze_c}
    \endgroup
    \label{fig:posevsgaze}
\end{figure*}

\paragraph{Richer prompting does not lead to consistent improvements.} \autoref{fig:accuracy_sport} shows that Cue-Directed prompting (explicitly instructing models to attend to player gaze and orientation) yields some improvement over Base prompting in some cases. However, these gains are inconsistent and don't close the large gap with human performance. Interestingly, performance sometimes degrades under Chain-of-Thought prompting compared to both Base and Cue-Directed prompting (GPT in soccer and basketball). Moreover, there are no clear overarching patterns of CoT effectiveness across models: while Gemini performs best when prompted using Chain-of-Thought in soccer, the model performs worst using the same prompting in basketball. These results suggest that the models have fundamental limitations in social understanding, in that they fail to use the relevant information even when it's explicitly pointed out to them.

\begin{figure*}[t]
        \centering
    \includegraphics[width=.85\linewidth]{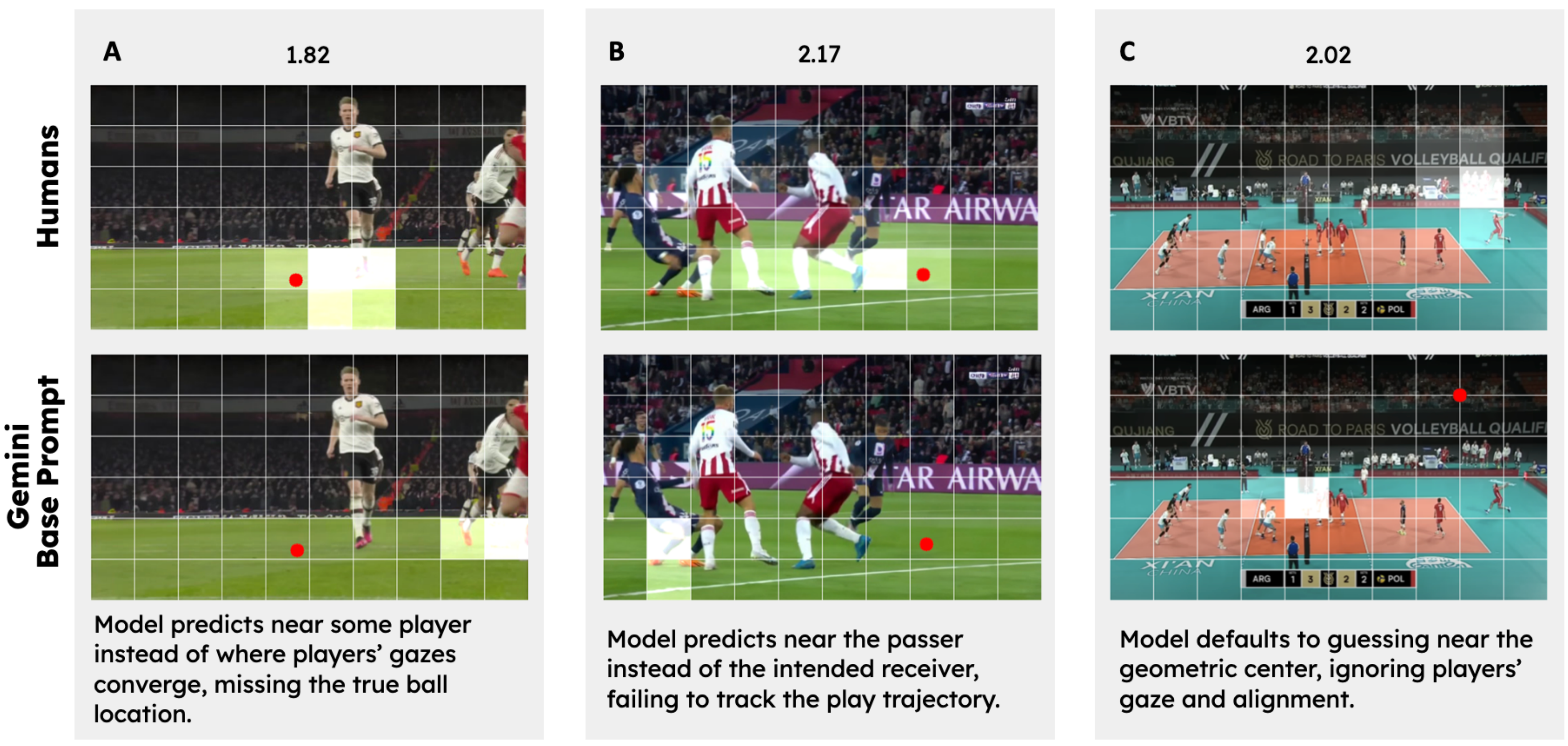}
    \caption{\textbf{Qualitative failure modes} demonstrated using Gemini using Base Prompt and human predictions. Red dots mark ground-truth ball locations, white squares mark model-predicted cells, and shaded heatmaps show prediction density. Wasserstein distances between the model and human for each example are shown in the center above each panel.}
    \begingroup
    \phantomsubcaption
    \label{subfig:quals_a}
    \phantomsubcaption
    \label{subfig:quals_b}
    \phantomsubcaption
    \label{subfig:quals_c}
    \endgroup
    \label{fig:quals}
\end{figure*}

\subsection{Qualitative Performance}
We analyze the semantic content of model reasoning through embedding-based similarity comparison (excluding Chain-of-Thought reasoning from Qwen and Llama due to compute constraints). For each model's reasoning text, we generate embeddings using Google's Gemini embedding model (models/embedding-001) and compare them against predefined pose and gaze reasoning templates (see Appendix). We compute the average cosine similarity between each reasoning and all templates in each category, then classify reasoning as pose-aligned if it is more similar to pose templates than gaze templates, and gaze-aligned otherwise.

\paragraph{Models attend more to pose than gaze in textual reasoning.}
Models disproportionately rely on pose cues relative to gaze across all sports (\autoref{subfig:posevsgaze_b}). This imbalance highlights a preference for coarse, body-level orientation signals rather than fine-grained gaze information. While humans exploit both gaze and pose relatively evenly across all sports (\autoref{subfig:posevsgaze_a}) , especially in less structured cues (Base and Cue-Directed Prompts), models default to pose, which may partly explain their systematic under-performance in sports, such as volleyball where the low size of the players make the pose cues harder to extract.

\paragraph{CoT prompting leads to more attention on gaze, but doesn't improve accuracy.}
Chain-of-thought prompting shifts model behavior. Compared to direct prompting, CoT outputs refer to gaze cues more frequently (\autoref{subfig:posevsgaze_c}). This suggests that explicit reasoning steps help models distribute attention across multiple social cues rather than disproportionately relying on pose alone. However, this attention difference only affected the textual reasoning. When the model guesses are evaluated on accuracy and distributional similarity to humans, CoT does not yield consistent gains, and in some cases, it even degrades performance. While prompting helps models describe a more human-like reasoning process -- one that mentions both pose and gaze cues -- it doesn't lead to better predictions of where the ball is.

\paragraph{Repeated failure modes.}
\autoref{fig:quals} illustrates common failure modes, using Gemini under Base Prompt as an example:
\begin{enumerate}
    \item \textbf{Neglect of gaze.} 
    Models fail to incorporate gaze cues into their predictions (as evidenced earlier by the lack of gaze reasoning in the embedding analysis), even when such cues provide strong evidence of ball location. For example, in \autoref{subfig:quals_a}, the model places its guess near the players' feet off to the right rather than recognizing the central player's gaze which the human guesses seem to attend to. 
  
    \item \textbf{Role confusion.} Models frequently misidentify which player has possession or is about to act. Instead of reasoning about roles within the play, they often resort to simply guessing near a player, as reflected in the proximity to players noted in our metrics. And as in \autoref{subfig:quals_b}, when the player identified is incorrect the low entropy further exacerbates the impact on the accuracy. 
    
    \item \textbf{Default-to-center heuristic.} \autoref{subfig:quals_c} provides a clear example of center bias: the model places its prediction directly in the middle of the image, at the net. This is an unlikely location for the ball, since play would have already terminated if the volleyball had struck the net. The prediction reflects a tendency to default to the geometric center rather than incorporating contextual cues.
    
\end{enumerate}

%==================================================================

\section{Conclusion}
We introduced \stb, a benchmark evaluating how vision-language models infer hidden objects from social cues in sports scenes. Across soccer, volleyball, and basketball, humans consistently outperform models by substantial margins. Models exhibit systematic biases like collapsing toward central regions and player proximity, and prompting strategies fail to close this gap. This points to deeper limitations in how current architectures perceive and reason about social cues like pose and gaze \citep[see also][]{qin2025simple}. Progress may require integrating perceptual priors, temporal information, or architectures explicitly designed to capture agentive and relational dynamics.

The need to understand these limitations is increasingly urgent. As VLMs are used to interpret complex scenes in interactive, visually rich environments like in embodied AI, their ability to make inferences from human pose, gaze, and spatial structure becomes central to evaluating what kinds of reasoning they can or cannot perform. Indeed, \stb isolates a core component of visual social inference that humans use effortlessly but current models struggle with. By releasing our dataset, pipeline, and evaluation code, we aim to make these limitations visible and to support systematic progress on this capability. 

%==================================================================

\clearpage
{
    \small
    \bibliographystyle{ieeenat_fullname}
    \bibliography{main}

@String(CVPR= {IEEE Conf. Comput. Vis. Pattern Recog.})

@String(CVPRW= {IEEE Conf. Comput. Vis. Pattern Recog. Worksh.})

@String(CVPR  = {CVPR})

@String(CVPRW= {CVPRW})

@article{qin2025simple,
	author = {Qin, Wenshuo and Isik, Leyla},
	journal = {arXiv preprint arXiv:2511.03988},
	title = {Simple 3D Pose Features Support Human and Machine Social Scene Understanding},
	year = {2025}}

@article{wang2025modeling,
	title = {Modeling {Other} {Minds}: {A} {Computational} {Account} of {Social} {Cognition} and {Its} {Development}},
	issn = {2640-7922},
	shorttitle = {Modeling {Other} {Minds}},
	url = {https://www.annualreviews.org/content/journals/10.1146/annurev-devpsych-111323-112016},
	doi = {10.1146/annurev-devpsych-111323-112016},
	urldate = {2025-10-10},
	journal = {Annual Review of Developmental Psychology},
	author = {Wang, Zihan and Davis, Isaac and Jara-Ettinger, Julian},
	month = sep,
	year = {2025},
}

@article{tremoulet2000perception,
	title = {Perception of {Animacy} from the {Motion} of a {Single} {Object}},
	volume = {29},
	issn = {0301-0066},
	url = {https://doi.org/10.1068/p3101},
	doi = {10.1068/p3101},
	number = {8},
	urldate = {2025-10-12},
	journal = {Perception},
	author = {Tremoulet, Patrice D and Feldman, Jacob},
	month = aug,
	year = {2000},
	pages = {943--951},
}

@article{todorov2015social,
	title = {Social {Attributions} from {Faces}: {Determinants}, {Consequences}, {Accuracy}, and {Functional} {Significance}},
	volume = {66},
	issn = {0066-4308, 1545-2085},
	shorttitle = {Social {Attributions} from {Faces}},
	url = {https://www.annualreviews.org/doi/10.1146/annurev-psych-113011-143831},
	doi = {10.1146/annurev-psych-113011-143831},
	number = {1},
	urldate = {2025-10-12},
	journal = {Annual Review of Psychology},
	author = {Todorov, Alexander and Olivola, Christopher Y. and Dotsch, Ron and Mende-Siedlecki, Peter},
	month = jan,
	year = {2015},
	pages = {519--545},
}

@article{degelder2015perception,
	title = {The perception of emotion in body expressions},
	volume = {6},
	issn = {1939-5086},
	url = {https://onlinelibrary.wiley.com/doi/abs/10.1002/wcs.1335},
	doi = {10.1002/wcs.1335},
	number = {2},
	urldate = {2025-10-12},
	journal = {WIREs Cognitive Science},
	author = {de Gelder, B. and de Borst, A.w. and Watson, R.},
	year = {2015},
	pages = {149--158},
}

@article{kleinke1986gaze,
	title = {Gaze and eye contact: {A} research review},
	volume = {100},
	issn = {1939-1455},
	shorttitle = {Gaze and eye contact},
	doi = {10.1037/0033-2909.100.1.78},
	number = {1},
	journal = {Psychological Bulletin},
	author = {Kleinke, Chris L.},
	year = {1986},
	pages = {78--100},
}

@misc{bai2025qwen25vltechnicalreport,
      title={Qwen2.5-VL Technical Report}, 
      author={Shuai Bai and Keqin Chen and Xuejing Liu and Jialin Wang and Wenbin Ge and Sibo Song and Kai Dang and Peng Wang and Shijie Wang and Jun Tang and Humen Zhong and Yuanzhi Zhu and Mingkun Yang and Zhaohai Li and Jianqiang Wan and Pengfei Wang and Wei Ding and Zheren Fu and Yiheng Xu and Jiabo Ye and Xi Zhang and Tianbao Xie and Zesen Cheng and Hang Zhang and Zhibo Yang and Haiyang Xu and Junyang Lin},
      year={2025},
      eprint={2502.13923},
      archivePrefix={arXiv},
      primaryClass={cs.CV},
      url={https://arxiv.org/abs/2502.13923}, 
}

@book{dennett1989intentional,
	title = {The {Intentional} {Stance}},
	isbn = {978-0-262-54053-7},
	language = {en-US},
	publisher = {The MIT Press},
	author = {Dennett, Daniel C.},
	month = mar,
	year = {1989},
}

@book{wellman2014making,
	address = {New York, NY, US},
	series = {Making minds: {How} theory of mind develops},
	title = {Making minds: {How} theory of mind develops},
	isbn = {978-0-19-933491-9},
	shorttitle = {Making minds},
	publisher = {Oxford University Press},
	author = {Wellman, Henry M.},
	year = {2014},
}

@article{sodian2011theory,
    title = {Theory of {{Mind}} in {{Infancy}}},
    author = {Sodian, Beate},
    year = {2011},
    journal = {Child Development Perspectives},
    volume = {5},
    number = {1},
    pages = {39--43},
    issn = {1750-8606},
    doi = {10.1111/j.1750-8606.2010.00152.x},
    langid = {english}
}

@article{heider1944experimental,
    title = {An {Experimental} {Study} of {Apparent} {Behavior}},
    volume = {57},
    issn = {0002-9556},
    url = {https://www.jstor.org/stable/1416950},
    doi = {10.2307/1416950},
    number = {2},
    urldate = {2022-06-17},
    journal = {The American Journal of Psychology},
    author = {Heider, Fritz and Simmel, Marianne},
    year = {1944},
    pages = {243--259},
}

@misc{pothiraj2025,
      title={CAPTURe: Evaluating Spatial Reasoning in Vision Language Models via Occluded Object Counting}, 
      author={Atin Pothiraj and Elias Stengel-Eskin and Jaemin Cho and Mohit Bansal},
      year={2025},
      eprint={2504.15485},
      archivePrefix={arXiv},
      primaryClass={cs.CV},
      url={https://arxiv.org/abs/2504.15485}, 
}

@inproceedings{li2024topviewrsvisionlanguagemodelstopview,
      title={TopViewRS: Vision-Language Models as Top-View Spatial Reasoners}, 
      author={Chengzu Li and Caiqi Zhang and Han Zhou and Nigel Collier and Anna Korhonen and Ivan Vulić},
      year={2024},
      booktitle={Proceedings of the 2024 Conference on Empirical Methods in Natural Language Processing},
      pages={1786–1807}
}

@misc{chen2024spatialvlmendowingvisionlanguagemodels,
      title={SpatialVLM: Endowing Vision-Language Models with Spatial Reasoning Capabilities}, 
      author={Boyuan Chen and Zhuo Xu and Sean Kirmani and Brian Ichter and Danny Driess and Pete Florence and Dorsa Sadigh and Leonidas Guibas and Fei Xia},
      year={2024},
      eprint={2401.12168},
      archivePrefix={arXiv},
      primaryClass={cs.CV},
      url={https://arxiv.org/abs/2401.12168}, 
}

@misc{radford2021learningtransferablevisualmodels,
      title={Learning Transferable Visual Models From Natural Language Supervision}, 
      author={Alec Radford and Jong Wook Kim and Chris Hallacy and Aditya Ramesh and Gabriel Goh and Sandhini Agarwal and Girish Sastry and Amanda Askell and Pamela Mishkin and Jack Clark and Gretchen Krueger and Ilya Sutskever},
      year={2021},
      eprint={2103.00020},
      archivePrefix={arXiv},
      primaryClass={cs.CV},
      url={https://arxiv.org/abs/2103.00020}, 
}

@article{byom2013theory,
  title   = {Theory of mind: mechanisms, methods, and new directions},
  author  = {Byom, Lindsey J. and Mutlu, Bilge},
  journal = {Frontiers in Human Neuroscience},
  volume  = {7},
  pages   = {413},
  year    = {2013},
  publisher = {Frontiers},
  doi     = {10.3389/fnhum.2013.00413},
  pmid    = {23964218},
  pmcid   = {PMC3737477}
}

@article{BaronCohen1985,
  author    = {Baron-Cohen, Simon and Leslie, Alan M. and Frith, Uta},
  title     = {Does the autistic child have a "theory of mind"?},
  journal   = {Cognition},
  year      = {1985},
  volume    = {21},
  number    = {1},
  pages     = {37--46},
  doi       = {10.1016/0010-0277(85)90022-8},
  issn      = {0010-0277},
  pmid      = {2934210},
  month     = oct
}

@article{FletcherWatson2014,
  author    = {Fletcher-Watson, Sue and McConnell, Fiona and Manola, Eleni and McConachie, Helen},
  title     = {Interventions based on the Theory of Mind cognitive model for autism spectrum disorder (ASD)},
  journal   = {Cochrane Database of Systematic Reviews},
  year      = {2014},
  volume    = {2014},
  number    = {3},
  pages     = {CD008785},
  doi       = {10.1002/14651858.CD008785.pub2},
  pmid      = {24652601},
  pmcid     = {PMC6923148},
  month     = mar
}

@article{TagerFlusberg2007,
  author    = {Tager-Flusberg, Helen},
  title     = {Evaluating the theory-of-mind hypothesis of autism},
  journal   = {Current Directions in Psychological Science},
  year      = {2007},
  volume    = {16},
  number    = {6},
  pages     = {311--315},
  doi       = {10.1111/j.1467-8721.2007.00527.x}
}

@misc{liu2025visiblebenchmarkingocclusionperception,
      title={Beyond the Visible: Benchmarking Occlusion Perception in Multimodal Large Language Models}, 
      author={Zhaochen Liu and Kaiwen Gao and Shuyi Liang and Bin Xiao and Limeng Qiao and Lin Ma and Tingting Jiang},
      year={2025},
      eprint={2508.04059},
      archivePrefix={arXiv},
      primaryClass={cs.CV},
      url={https://arxiv.org/abs/2508.04059}, 
}

@misc{sap2019socialiqacommonsensereasoningsocial,
      title={SocialIQA: Commonsense Reasoning about Social Interactions}, 
      author={Maarten Sap and Hannah Rashkin and Derek Chen and Ronan LeBras and Yejin Choi},
      year={2019},
      eprint={1904.09728},
      archivePrefix={arXiv},
      primaryClass={cs.CL},
      url={https://arxiv.org/abs/1904.09728}, 
}

@misc{zellers2019recognitioncognitionvisualcommonsense,
      title={From Recognition to Cognition: Visual Commonsense Reasoning}, 
      author={Rowan Zellers and Yonatan Bisk and Ali Farhadi and Yejin Choi},
      year={2019},
      eprint={1811.10830},
      archivePrefix={arXiv},
      primaryClass={cs.CV},
      url={https://arxiv.org/abs/1811.10830}, 
}

@misc{mathur2025socialgenomegroundedsocial,
      title={Social Genome: Grounded Social Reasoning Abilities of Multimodal Models}, 
      author={Leena Mathur and Marian Qian and Paul Pu Liang and Louis-Philippe Morency},
      year={2025},
      eprint={2502.15109},
      archivePrefix={arXiv},
      primaryClass={cs.CL},
      url={https://arxiv.org/abs/2502.15109}, 
}

@misc{xu2025socialmazebenchmarkevaluatingsocial,
      title={SocialMaze: A Benchmark for Evaluating Social Reasoning in Large Language Models}, 
      author={Zixiang Xu and Yanbo Wang and Yue Huang and Jiayi Ye and Haomin Zhuang and Zirui Song and Lang Gao and Chenxi Wang and Zhaorun Chen and Yujun Zhou and Sixian Li and Wang Pan and Yue Zhao and Jieyu Zhao and Xiangliang Zhang and Xiuying Chen},
      year={2025},
      eprint={2505.23713},
      archivePrefix={arXiv},
      primaryClass={cs.CL},
      url={https://arxiv.org/abs/2505.23713}, 
}

@inproceedings{hitom,
    title = "Hi-{T}o{M}: A Benchmark for Evaluating Higher-Order Theory of Mind Reasoning in Large Language Models",
    author = "Wu, Yufan  and
      He, Yinghui  and
      Jia, Yilin  and
      Mihalcea, Rada  and
      Chen, Yulong  and
      Deng, Naihao",
    editor = "Bouamor, Houda  and
      Pino, Juan  and
      Bali, Kalika",
    booktitle = "Findings of the Association for Computational Linguistics: EMNLP 2023",
    month = dec,
    year = "2023",
    address = "Singapore",
    publisher = "Association for Computational Linguistics",
    url = "https://aclanthology.org/2023.findings-emnlp.717/",
    doi = "10.18653/v1/2023.findings-emnlp.717",
    pages = "10691--10706"
}

@misc{tomchallenges,
      title={ToMChallenges: A Principle-Guided Dataset and Diverse Evaluation Tasks for Exploring Theory of Mind}, 
      author={Xiaomeng Ma and Lingyu Gao and Qihui Xu},
      year={2024},
      eprint={2305.15068},
      archivePrefix={arXiv},
      primaryClass={cs.CL},
      url={https://arxiv.org/abs/2305.15068}, 
}

@inproceedings{
gandhi2023understanding,
title={Understanding Social Reasoning in Language Models with Language Models},
author={Kanishk Gandhi and Jan-Philipp Fr{\"a}nken and Tobias Gerstenberg and Noah Goodman},
booktitle={Thirty-seventh Conference on Neural Information Processing Systems Datasets and Benchmarks Track},
year={2023},
url={https://openreview.net/forum?id=8bqjirgxQM}
}

@misc{chen2024emotionqueenbenchmarkevaluatingempathy,
      title={EmotionQueen: A Benchmark for Evaluating Empathy of Large Language Models}, 
      author={Yuyan Chen and Hao Wang and Songzhou Yan and Sijia Liu and Yueze Li and Yi Zhao and Yanghua Xiao},
      year={2024},
      eprint={2409.13359},
      archivePrefix={arXiv},
      primaryClass={cs.CL},
      url={https://arxiv.org/abs/2409.13359}, 
}

@inproceedings{apathy,
 author = {Huang, Jen-tse and Lam, Man Ho and Li, Eric John and Ren, Shujie and Wang, Wenxuan and Jiao, Wenxiang and Tu, Zhaopeng and Lyu, Michael R.},
 booktitle = {Advances in Neural Information Processing Systems},
 editor = {A. Globerson and L. Mackey and D. Belgrave and A. Fan and U. Paquet and J. Tomczak and C. Zhang},
 pages = {97053--97087},
 publisher = {Curran Associates, Inc.},
 title = {Apathetic or Empathetic? Evaluating LLMs\textquotesingle  Emotional Alignments with Humans},
 url = {https://proceedings.neurips.cc/paper_files/paper/2024/file/b0049c3f9c53fb06f674ae66c2cf2376-Paper-Conference.pdf},
 volume = {37},
 year = {2024}
}

@misc{moralchoice,
      title={Evaluating the Moral Beliefs Encoded in LLMs}, 
      author={Nino Scherrer and Claudia Shi and Amir Feder and David M. Blei},
      year={2023},
      eprint={2307.14324},
      archivePrefix={arXiv},
      primaryClass={cs.CL},
      url={https://arxiv.org/abs/2307.14324}, 
}

@misc{morables,
      title={MORABLES: A Benchmark for Assessing Abstract Moral Reasoning in LLMs with Fables}, 
      author={Matteo Marcuzzo and Alessandro Zangari and Andrea Albarelli and Jose Camacho-Collados and Mohammad Taher Pilehvar},
      year={2025},
      eprint={2509.12371},
      archivePrefix={arXiv},
      primaryClass={cs.CL},
      url={https://arxiv.org/abs/2509.12371}, 
}

@misc{drex,
      title={D-REX: A Benchmark for Detecting Deceptive Reasoning in Large Language Models}, 
      author={Satyapriya Krishna and Andy Zou and Rahul Gupta and Eliot Krzysztof Jones and Nick Winter and Dan Hendrycks and J. Zico Kolter and Matt Fredrikson and Spyros Matsoukas},
      year={2025},
      eprint={2509.17938},
      archivePrefix={arXiv},
      primaryClass={cs.CL},
      url={https://arxiv.org/abs/2509.17938}, 
}

@misc{chen2024deceptiondetectiondeeperdataset,
      title={Can Deception Detection Go Deeper? Dataset, Evaluation, and Benchmark for Deception Reasoning}, 
      author={Kang Chen and Zheng Lian and Haiyang Sun and Rui Liu and Jiangyan Yi and Bin Liu and Jianhua Tao},
      year={2024},
      eprint={2402.11432},
      archivePrefix={arXiv},
      primaryClass={cs.CL},
      url={https://arxiv.org/abs/2402.11432}, 
}

@inproceedings{negotiation,
 author = {Abdelnabi, Sahar and Gomaa, Amr and Sivaprasad, Sarath and Sch\"{o}nherr, Lea and Fritz, Mario},
 booktitle = {Advances in Neural Information Processing Systems},
 editor = {A. Globerson and L. Mackey and D. Belgrave and A. Fan and U. Paquet and J. Tomczak and C. Zhang},
 pages = {83548--83599},
 publisher = {Curran Associates, Inc.},
 title = {Cooperation, Competition, and Maliciousness: LLM-Stakeholders Interactive Negotiation},
 url = {https://proceedings.neurips.cc/paper_files/paper/2024/file/984dd3db213db2d1454a163b65b84d08-Paper-Datasets_and_Benchmarks_Track.pdf},
 volume = {37},
 year = {2024}
}

@inproceedings{mathur-etal-2024-advancing,
    title = "Advancing Social Intelligence in {AI} Agents: Technical Challenges and Open Questions",
    author = "Mathur, Leena  and
      Liang, Paul Pu  and
      Morency, Louis-Philippe",
    editor = "Al-Onaizan, Yaser  and
      Bansal, Mohit  and
      Chen, Yun-Nung",
    booktitle = "Proceedings of the 2024 Conference on Empirical Methods in Natural Language Processing",
    month = nov,
    year = "2024",
    url = "https://aclanthology.org/2024.emnlp-main.1143/",
    doi = "10.18653/v1/2024.emnlp-main.1143",
    pages = "20541--20560"
}

@INPROCEEDINGS{socialiq1,
  author={Zadeh, Amir and Chan, Michael and Liang, Paul Pu and Tong, Edmund and Morency, Louis-Philippe},
  booktitle={2019 IEEE/CVF Conference on Computer Vision and Pattern Recognition (CVPR)}, 
  title={Social-IQ: A Question Answering Benchmark for Artificial Social Intelligence}, 
  year={2019},
  volume={},
  number={},
  pages={8799-8809},
  doi={10.1109/CVPR.2019.00901}}

@article{wilf2023social,
  title={Social-iq 2.0 challenge: Benchmarking multimodal social understanding},
  author={Wilf, Alex and Mathur, Leena and Mathew, Sheryl and Ko, Claire and Kebe, Youssouf and Liang, Paul Pu and Morency, Louis-Philippe},
  year={2023}
}

@misc{mmtomqa,
      title={MMToM-QA: Multimodal Theory of Mind Question Answering}, 
      author={Chuanyang Jin and Yutong Wu and Jing Cao and Jiannan Xiang and Yen-Ling Kuo and Zhiting Hu and Tomer Ullman and Antonio Torralba and Joshua B. Tenenbaum and Tianmin Shu},
      year={2024},
      eprint={2401.08743},
      archivePrefix={arXiv},
      primaryClass={cs.AI},
      url={https://arxiv.org/abs/2401.08743}, 
}

@misc{mumatom,
      title={MuMA-ToM: Multi-modal Multi-Agent Theory of Mind}, 
      author={Haojun Shi and Suyu Ye and Xinyu Fang and Chuanyang Jin and Leyla Isik and Yen-Ling Kuo and Tianmin Shu},
      year={2025},
      eprint={2408.12574},
      archivePrefix={arXiv},
      primaryClass={cs.AI},
      url={https://arxiv.org/abs/2408.12574}, 
}

@misc{li2025egotombenchmarkingtheorymind,
      title={EgoToM: Benchmarking Theory of Mind Reasoning from Egocentric Videos}, 
      author={Yuxuan Li and Vijay Veerabadran and Michael L. Iuzzolino and Brett D. Roads and Asli Celikyilmaz and Karl Ridgeway},
      year={2025},
      eprint={2503.22152},
      archivePrefix={arXiv},
      primaryClass={cs.CV},
      url={https://arxiv.org/abs/2503.22152}, 
}

@misc{fan2025somitomevaluatingmultiperspectivetheory,
      title={SoMi-ToM: Evaluating Multi-Perspective Theory of Mind in Embodied Social Interactions}, 
      author={Xianzhe Fan and Xuhui Zhou and Chuanyang Jin and Kolby Nottingham and Hao Zhu and Maarten Sap},
      year={2025},
      eprint={2506.23046},
      archivePrefix={arXiv},
      primaryClass={cs.CL},
      url={https://arxiv.org/abs/2506.23046}, 
}

@misc{xiao2021nextqanextphasequestionansweringexplaining,
      title={NExT-QA:Next Phase of Question-Answering to Explaining Temporal Actions}, 
      author={Junbin Xiao and Xindi Shang and Angela Yao and Tat-Seng Chua},
      year={2021},
      eprint={2105.08276},
      archivePrefix={arXiv},
      primaryClass={cs.CV},
      url={https://arxiv.org/abs/2105.08276}, 
}

@misc{wu2023acquireddatasetansweringcounterfactual,
      title={ACQUIRED: A Dataset for Answering Counterfactual Questions In Real-Life Videos}, 
      author={Te-Lin Wu and Zi-Yi Dou and Qingyuan Hu and Yu Hou and Nischal Reddy Chandra and Marjorie Freedman and Ralph M. Weischedel and Nanyun Peng},
      year={2023},
      eprint={2311.01620},
      archivePrefix={arXiv},
      primaryClass={cs.CV},
      url={https://arxiv.org/abs/2311.01620}, 
}

@ARTICLE{affectnet,
  author={Mollahosseini, Ali and Hasani, Behzad and Mahoor, Mohammad H.},
  journal={IEEE Transactions on Affective Computing}, 
  title={AffectNet: A Database for Facial Expression, Valence, and Arousal Computing in the Wild}, 
  year={2019},
  volume={10},
  number={1},
  pages={18-31},
  doi={10.1109/TAFFC.2017.2740923}}

@INPROCEEDINGS{emotic,
  author={Kosti, Ronak and Alvarez, Jose M. and Recasens, Adria and Lapedriza, Agata},
  booktitle={2017 IEEE Conference on Computer Vision and Pattern Recognition Workshops (CVPRW)}, 
  title={EMOTIC: Emotions in Context Dataset}, 
  year={2017},
  volume={},
  number={},
  pages={2309-2317},
  doi={10.1109/CVPRW.2017.285}}

@misc{li2018visualsocialrelationshiprecognition,
      title={Visual Social Relationship Recognition}, 
      author={Junnan Li and Yongkang Wong and Qi Zhao and Mohan S. Kankanhalli},
      year={2018},
      eprint={1812.05917},
      archivePrefix={arXiv},
      primaryClass={cs.CV},
      url={https://arxiv.org/abs/1812.05917}, 
}

@article{opencv_library,
    author = {Bradski, G.},
    journal = {Dr. Dobb's Journal of Software Tools},
    title = {{The OpenCV Library}},
    year = {2000}
}

@software{yolov8_ultralytics,
  author = {Glenn Jocher and Ayush Chaurasia and Jing Qiu},
  title = {Ultralytics YOLOv8},
  version = {8.0.0},
  year = {2023},
  url = {https://github.com/ultralytics/ultralytics},
  orcid = {0000-0001-5950-6979, 0000-0002-7603-6750, 0000-0003-3783-7069},
  license = {AGPL-3.0}
}
}

% WARNING: do not forget to delete the supplementary pages from your submission 
\clearpage
\setcounter{page}{1}
\maketitlesupplementary

\section{System Prompts}
\paragraph{Base Prompt.} Tells the model and human that the ball is removed and asks for a guess of where it would've been 
\begin{lstlisting}[language=Python, basicstyle=\ttfamily\scriptsize]
    The ball has been removed from this {sport} image. Your task is to infer the most likely location of the ball.
    Respond in the following format:
    Reasoning: <Explain where the ball is likely located and why.>
    Cell: <What grid cell is the ball most likely located in? Respond with a label like F4.> 
\end{lstlisting}
\paragraph{Cue-Directed Prompt} In addition to Base Prompt, tells the model to consider player gaze, pose, and positions when predicting.

\begin{lstlisting}[language=Python, basicstyle=\ttfamily\scriptsize]
    The ball has been removed from this {sport} image. Your task is to infer the most likely location of the ball.
    The location of the players, where they are looking and their positions can help you infer the location of the ball.
    Respond in the following format:
    Reasoning: <Explain where the ball is likely located and why.>
    Cell: <What grid cell is the ball most likely located in? Respond with a label like F4.>
\end{lstlisting}

\paragraph{Chain-of-Thought Prompt} Contains an added intermediate social reasoning step before making the final prediction. 
\begin{enumerate}
    \item The model is asked to answer three questions to extract relevant visual information from the scene
    
\begin{lstlisting}[language=Python, basicstyle=\ttfamily\scriptsize]
    The ball has been removed from this {sport} image. Your task is to infer the most likely location of the ball.
    The location of the players, where they are looking and their positions can help you infer the location of the ball.
    Respond in the following format:
    Reasoning: <Explain where the ball is likely located and why.>
    Cell: <What grid cell is the ball most likely located in? Respond with a label like F4.>
\end{lstlisting}
    \item The model then receives both the original instruction (that the ball has been removed) and the {context} observations
\begin{lstlisting}[language=Python, basicstyle=\ttfamily\scriptsize]
    The ball has been removed from this {sport} image. Here are some observations:
    {context}
    The above information could help you infer the ball's location.
    Respond in the following format:
    Reasoning: <Explain where the ball is likely located and why.>
    Cell: <What grid cell is the ball most likely located in? Respond with a label like F4.>
\end{lstlisting}
\end{enumerate}

\section{Evaluation Metrics Implementation}
\subsection{Wasserstein Distance} 
\begin{figure}
    \centering
    \includegraphics[width=1\linewidth]{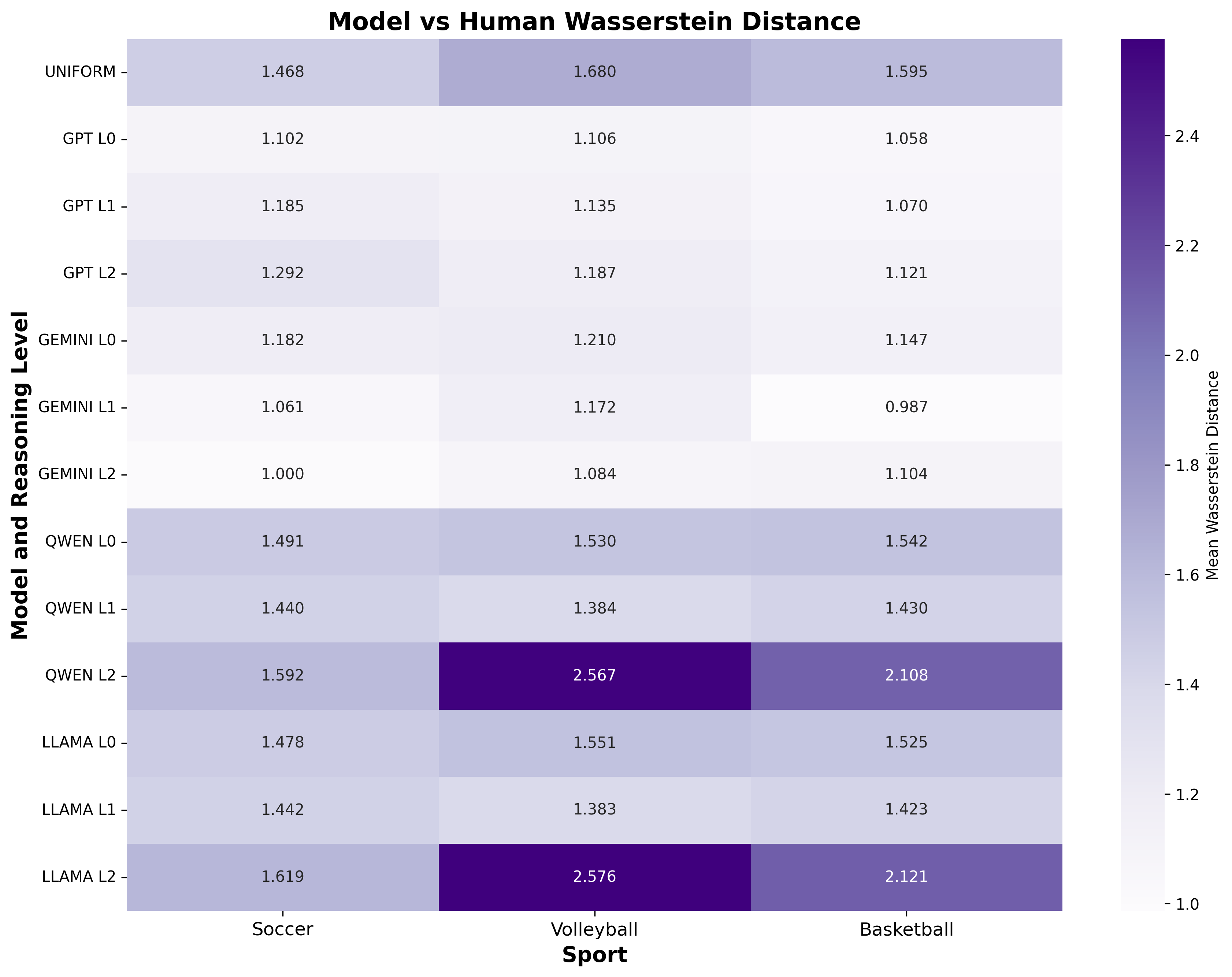}
    \caption{Wasserstein Distances with all the levels and models. L0 refers to Base prompt, L1 to cue-directed prompt and L2 to chain-of-thought prompt.}
    \label{fig:bigwasserstein}
\end{figure}

We used \texttt{wasserstein\_distance} function from the \texttt{scipy} packege with coordinate weights representing the probability mass at each grid cell to calculate the Wasserstein distances. The distances from each level of each model is shown in \autoref{fig:bigwasserstein}. 

\subsection{Player Proximity}
We detect players in the images via the \texttt{YOLOv8} ``person'' class per frame and then remove audience in the background through a lightweight manual pass. The review UI shows each image with proposed boxes; annotators uncheck any non-players. The size and location of these boxes are then relevant to the proximity analysis. 

\paragraph{Threshold fitting and robustness}
\begin{figure}[t]
    \centering
    \includegraphics[width=1\linewidth]{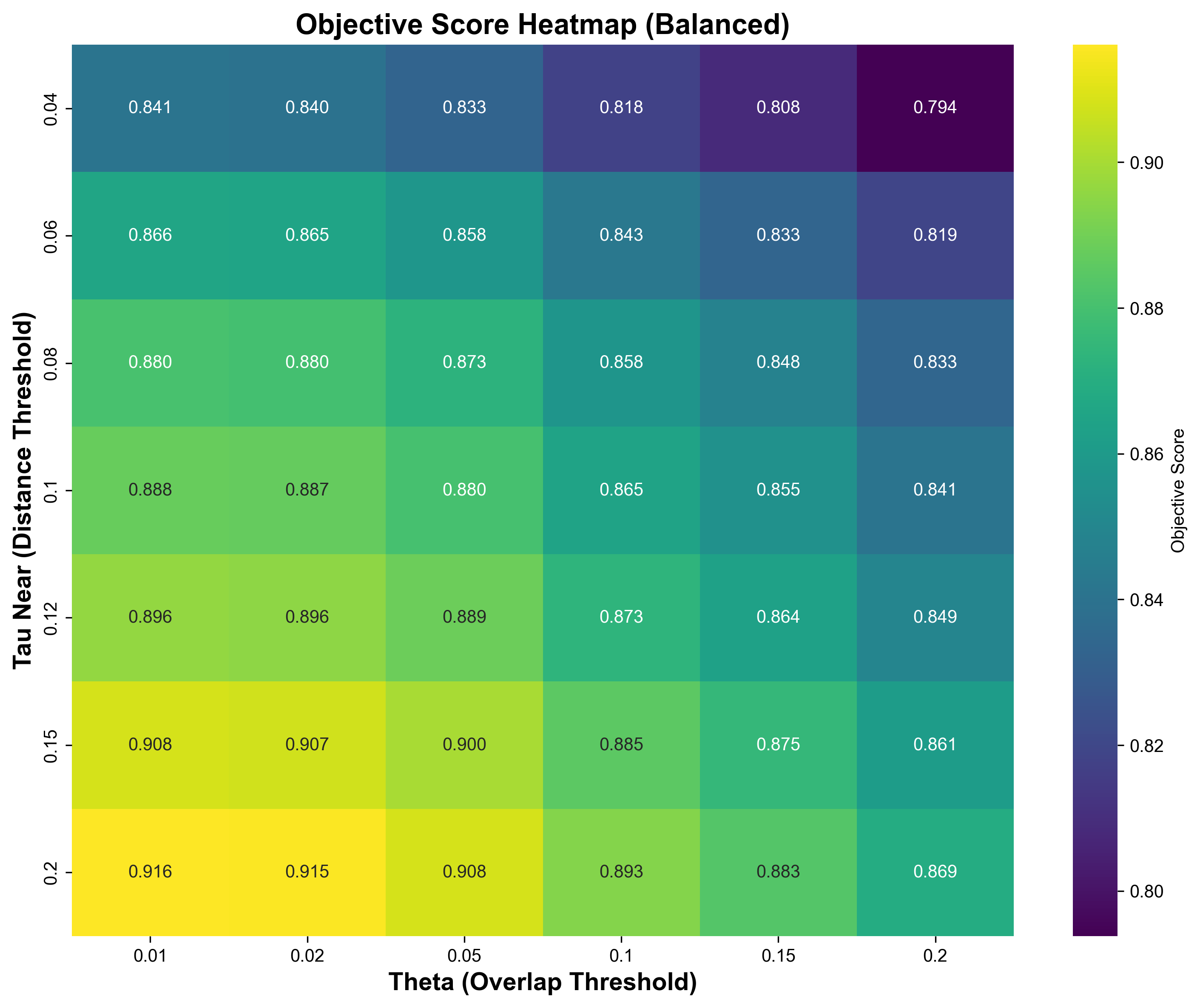}
    \caption{Balanced objective values over the grid of $\tau_{\text{near}}$ and $\theta$.
The objective peaks broadly around $\tau_{\text{near}}=0.08$ and $\theta=0.02$,
suggesting the chosen thresholds are stable and conservative.}
    \label{fig:threshold_heatmaps}
\end{figure}

\label{app:threshold_fitting}
We determine the thresholds for the Near‐Player Rate (NR) and Near‐Overlap Rate (OR) metrics through a grid search over 
\[
\tau_{\text{near}} \in [0.04, 0.20], \quad \theta \in [0.01, 0.20],
\]
where $\tau_{\text{near}}$ represents the distance threshold as a fraction of the image diagonal, and $\theta$ denotes the minimum overlap fraction between a predicted grid cell and any player box.

To guide selection, we define a \emph{balanced objective} that equally weights NR and OR:
\[
\mathcal{O}(\tau,\theta) = \tfrac{1}{2} \bigl[\text{NR}(\tau,\theta) + \text{OR}(\tau,\theta)\bigr].
\]
This formulation treats proximity to players (NR) and geometric overlap (OR) as complementary aspects of spatial behavior, ensuring thresholds that capture both close and intersecting predictions without overfitting to one metric.

The balanced objective exhibited a broad plateau centered near
\[
\tau_{\text{near}} = 0.08, \quad \theta = 0.02,
\]
with neighboring configurations yielding comparable scores (within~1\% of the maximum; Fig.~\ref{fig:threshold_heatmaps}). 
Empirical distributions revealed that most predicted or true ball regions already overlap with player boxes (median normalized distance $=0$; median overlap fraction $\approx 0.56$), suggesting that lower thresholds would saturate the metrics. 
The selected values therefore represent conservative, fitted cutoffs that maintain discriminability and stability across models and sports.

\section{Example model outputs}
\label{sec:appendix-outputs}
\begin{figure}[t!]
    \centering
    \includegraphics[width=1\linewidth]{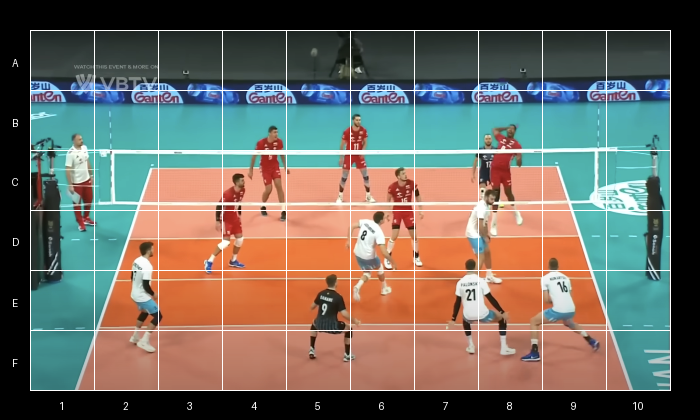}
    \caption{An example of the volleyball image shown to the models. The ball is masked, and a grid is overlaid; the responses in the following tables are generated based on this image.}
    \label{fig:volleyballexample}
\end{figure}

We show an example image from our dataset with the volleyball masked and grid overlaid, followed by model responses at each of the three prompting levels from Gemini and GPT (\autoref{fig:volleyballexample}). These examples illustrate how models interpret player positions, gaze, and team configurations to localize the ball, and how their reasoning changes with increasing prompt information.

\begin{table*}[t]
\centering
\caption{Example GPT output on the volleyball image (Base Prompt).}
\label{tab:example-d6}
\renewcommand{\arraystretch}{1.3}
\begin{tabularx}{\linewidth}{>{\columncolor{gray!15}\bfseries}p{0.22\linewidth} X}
\toprule
Reasoning &
The player in the red jersey on the right side of the net is in the air with his arm extended, appearing to be in the act of spiking or hitting the ball. The other players on the red team are looking towards the same area, indicating the ball is likely near his hand. The ball would be just above the net and close to this player's hand \\
\midrule
Grid cell guess & \texttt{Cell: C9} \\
\bottomrule
\end{tabularx}
\end{table*}
\begin{table*}[h]
\centering
\caption{Example Gemini output on the volleyball image (Base Prompt).}
\label{tab:example-d6b}
\renewcommand{\arraystretch}{1.3}
\begin{tabularx}{\linewidth}{>{\columncolor{gray!15}\bfseries}p{0.22\linewidth} X}
\toprule
Reasoning &
The player in cell D4 is in the middle of spiking the volleyball. His arm is extended upward and towards the net. Therefore, the ball is most likely in that vicinity as well. \\
\midrule
Grid cell guess & \texttt{Cell: D4} \\
\bottomrule
\end{tabularx}

\end{table*}
\begin{table*}[t]
\centering
\caption{Example GPT output on a volleyball image (Cue-Directed Prompt).}
\label{tab:example-d6}
\renewcommand{\arraystretch}{1.3}
\begin{tabularx}{\linewidth}{>{\columncolor{gray!15}\bfseries}p{0.22\linewidth} X}
\toprule
Reasoning &
The players in white are focused and positioned as if they are about to receive or set the ball. The player in white at cell E8 is looking upwards and has his hands raised, indicating he is preparing to make contact with the ball. The players in red on the opposite side are also looking towards the same area, suggesting the ball is above the net near the middle of the court. Given the positioning and gaze of the players, the ball is most likely just above the net around the center-right area. \\
\midrule
Grid cell guess & \texttt{Cell: C7} \\
\bottomrule
\end{tabularx}
\end{table*}

\begin{table*}[t]
\centering
\caption{Example Gemini output on a volleyball image (Cue-Directed Prompt).}
\label{tab:example-d6}
\renewcommand{\arraystretch}{1.3}
\begin{tabularx}{\linewidth}{>{\columncolor{gray!15}\bfseries}p{0.22\linewidth} X}
\toprule
Reasoning &
The player at position 6 in the D row is clearly hitting the ball. Given that the player at the front net at the C row is looking down as well as the players in the bottom right, it would be likely the ball is in the air after the player at position 6 in the D row is hitting it. \\
\midrule
Grid cell guess & \texttt{Cell: D6} \\
\bottomrule
\end{tabularx}
\end{table*}

\begin{table*}[t]
\centering
\caption{Example GPT output for the volleyball image (Chain-of-Thought Prompt).}
\label{tab:example-output}
\renewcommand{\arraystretch}{1.3}
\begin{tabularx}{\linewidth}{>{\columncolor{gray!15}\bfseries}p{0.25\linewidth} p{0.7\linewidth}}
\toprule
\textbf{Where are the players located?} & 
The players are located at the following grid positions: \newline
- Player in white jersey near the left side: E2 \newline
- Player in white jersey near the center: E6 \newline
- Player in white jersey near the right side: E8 \newline
- Player in black jersey near the center: E5 \newline
- Player in black jersey near the right side: E7 \newline
- Player in red jersey near the left side: C3 \newline
- Player in red jersey near the center-left: C5 \newline
- Player in red jersey near the center: C6 \newline
- Player in red jersey near the center-right: C7 \newline
- Player in red jersey near the right side: C8 \\

\midrule
Where are the players looking? &
The players are looking in the following directions: \newline
- The player in white at E2 is looking towards the net, slightly to the right. \newline
- The player in white at E6 is looking upwards and slightly to the right. \newline
- The player in white at E7 is looking upwards and slightly to the left. \newline
- The player in white at E9 is looking upwards and slightly to the left. \newline
- The player in black at E5 is looking upwards and slightly to the right. \newline
- The player in red at C4 is looking upwards and slightly to the right. \newline
- The player in red at C6 is looking upwards and slightly to the left. \newline
- The player in red at C7 is looking upwards and slightly to the left. \newline
- The player in red at C8 is looking upwards and slightly to the left. \newline
- The player in red at B9 is looking upwards and slightly to the left. \newline
\newline
Overall, most players are looking upwards, likely tracking the volleyball in play. \\

\midrule
How are the players positioned? &
The players are positioned as follows: \newline
\newline
\textit{Team in red (top side of the net):} \newline
- One player at B4 \newline
- One player at B6 \newline
- One player at B8 \newline
- One player at C3 \newline
- One player at C5 \newline
- One player at C7 \newline
\newline
\textit{Team in white and blue (bottom side of the net):} \newline
- One player at E2 \newline
- One player at E6 \newline
- One player at E7 \newline
- One player at E8 \newline
- One player at F5 \newline
- One player at F6 \\

\midrule
Reasoning &
Most players are looking upwards, indicating the ball is in the air. The players in white and black near the center (E5, E6, E7) are looking upwards and slightly to the right or left, suggesting the ball is above or near the net. The red team players near the net (C5, C6, C7, C8) are also looking upwards and slightly to the left, indicating the ball is likely above the net or just past it on the white team's side. Given the positioning and gaze directions, the ball is most likely just above the net around the center, near the grid cell between E5 and F5. \\

\midrule
Grid cell guess & \texttt{E5} \\
\bottomrule
\end{tabularx}
\end{table*}

\begin{table*}[h]
\centering
\caption{Example Gemini output on the volleyball image (Chain-of-Thought Prompt).}
\label{tab:gemini-output}
\renewcommand{\arraystretch}{1.3}
\begin{tabularx}{\linewidth}{>{\columncolor{gray!15}\bfseries}p{0.22\linewidth} X}
\toprule
Where are the players located? &
Here are the locations of the players in the image: \newline
* **1, C**: Player in white and red. \newline
* **2, D**: Player in white and blue. \newline
* **3, D**: Player in red. \newline
* **4, C**: Player in red. \newline
* **5, E**: Player in dark blue and white. \newline
* **5, B**: Player in red. \newline
* **6, D**: Player in white and blue. \newline
* **6, C**: Player in red. \newline
* **7, E**: Player numbered 21 in white and blue. \newline
* **8, C**: Player in red. \newline
* **8, D**: Player in white and blue. \newline
* **9, E**: Player numbered 16 in white and blue. \newline
* **9, B**: Player in red. \newline
* **10, C**: Player in white and blue. \\
\midrule
Where are the players looking? &
Here's an analysis of where the players are likely looking in the image, keeping in mind that it's difficult to be 100\% certain without seeing their eyes clearly: \newline
**Players on the left side (likely about to receive the ball):** They are likely looking at the ball or the opposing players about to hit the ball. Some of them might be tracking the anticipated trajectory of the ball. \newline
**Players on the right side (likely having just hit the ball):** They are most likely watching the ball after hitting it and looking at the opposing team to see how the receiving team will handle it. \\
\midrule
How are the players positioned? &
Here is the approximate positioning of the players based on the grid: \newline
\newline
**Team 1 (Left Side):** \newline
* Player 1: A-1/2 \newline
* Player 2: D-2/3 \newline
* Player 3: C-4 \newline
* Player 4: D-6 \newline
\newline
**Team 2 (Right Side):** \newline
* Player 1: B-9 \newline
* Player 2: D-8 \newline
* Player 3: E-7 \newline
* Player 4: F-9 \newline
\newline
**Mid-Court Players:** \newline
* Player 5: E-5 \newline
* Player 6: B-5 \newline
* Player 7: C-6 \newline
* Player 8: B-8 \newline
* Player 9: C-4 \\
\midrule
Reasoning &
Considering the positions of the players on both sides of the net and how they are looking, it appears the ball was hit from the right side (Team 2). Player E7 is watching the ball and is looking at it coming towards his side of the court. The players on the left side (Team 1) are getting ready to receive. With that knowledge I can assume that the ball is in the E4 grid position. \\
\midrule
Grid cell guess & \texttt{Cell: E4} \\
\bottomrule
\end{tabularx}
\end{table*}

\section{Code and Dataset}
All code required to reproduce the experiments, analyses, and figures is available at
[REDACTED]. 
The corresponding dataset, including evaluation splits and annotation metadata, can be accessed at 
[REDACTED].

\end{document}